\documentclass[review]{elsarticle}

\usepackage{latexsym}
\usepackage{graphicx}
\usepackage{amsfonts,amssymb,amsmath}
\usepackage{hyperref}
\usepackage{algorithmic}
\usepackage{textcomp}
\usepackage{xcolor}
\usepackage{import}
\usepackage{tabularx}
\usepackage{array}
\usepackage{tabu}
\usepackage{multirow}
\usepackage{siunitx}
\usepackage{booktabs}
\usepackage{times}
\usepackage{fancyhdr,graphicx,amsmath,amssymb}
\usepackage[ruled,vlined,linesnumbered]{algorithm2e}
\usepackage{mathtools}
\include{pythonlisting}
\usepackage{subcaption}
\usepackage{silence}
\usepackage{bbm}

\usepackage{amsthm}
\usepackage[utf8]{inputenc}
\usepackage[english]{babel}

\usepackage[ruled]{algorithm2e}
\usepackage{color,soul}

\SetAlFnt{\small}
\SetAlCapFnt{\small}
\SetAlCapNameFnt{\small}
\SetAlCapHSkip{0pt}
\IncMargin{-\parindent}
\usepackage{multirow}
\usepackage{algorithmic}
\usepackage{amsfonts}
\usepackage{ctable}
\usepackage{amsmath}
\usepackage{amsthm}
\usepackage{nth}
\usepackage{graphicx}
\usepackage{ragged2e}

\newcommand\Topspace{\rule{0pt}{4ex}}     
\newcommand\Bottomspace{\rule[-2ex]{0pt}{0pt}}  

\def\BibTeX{{\rm B\kern-.05em{\sc i\kern-.025em b}\kern-.08em T\kern-.1667em\lower.7ex\hbox{E}\kern-.125emX}}

\DeclarePairedDelimiter\abs{\lvert}{\rvert}%
\DeclarePairedDelimiter\norm{\lVert}{\rVert}%

\makeatletter
\renewcommand*\env@matrix[1][*\c@MaxMatrixCols c]{%
  \hskip -\arraycolsep
  \let\@ifnextchar\new@ifnextchar
  \array{#1}}
\let\oldabs\abs
\def\abs{\@ifstar{\oldabs}{\oldabs*}}
\let\oldnorm\norm
\def\norm{\@ifstar{\oldnorm}{\oldnorm*}}
\makeatother

\DeclareMathOperator*{\argmax}{argmax}

\begin{document}

\begin{frontmatter} 

\title{Joint Path planning and Power Allocation of a Cellular-Connected UAV using Apprenticeship Learning via Deep Inverse Reinforcement Learning}

\author[mymainaddress]{Alireza Shamsoshoara}
\ead{alireza\_shamsoshoara@nau.edu}
\author[mysecondaddress]{Fatemeh Lotfi}
\ead{flotfi@clemson.edu}
\author[mythirdaddress]{Sajad Mousavi}
\ead{seyedsajad\_mousavi@hms.harvard.edu}
\author[mysecondaddress]{Fatemeh Afghah}
\ead{fafghah@clemson.edu}
\author[myfourthaddress]{\.{I}smail G\"{u}ven\c{c}}
\ead{iguvenc@ncsu.edu}

\address[mymainaddress]{School of Informatics, Computing and Cyber Systems at Northern Arizona University, Flagstaff, AZ, USA. }
\address[mysecondaddress]{Department of Electrical and Computer Engineering, Clemson University, Clemson, SC, USA. }
\address[mythirdaddress]{Harvard Medical School, Boston, MA, USA. }
\address[myfourthaddress]{North Carolina State University, Raleigh, NC, USA. }






\begin{abstract}
This paper investigates an interference-aware joint path planning and power allocation mechanism for a cellular-connected unmanned aerial vehicle (UAV) in a sparse suburban environment. The UAV's goal is to fly from an initial point and reach a destination point by moving along the cells to guarantee the required quality of service (QoS). In particular, the UAV aims to maximize its uplink throughput and minimize the level of interference to the ground user equipment (UEs) connected to
the neighbor cellular BSs, considering the shortest path and flight resource limitation. Expert knowledge is used to experience the scenario and define the desired behavior for the sake of the agent (i.e., UAV) training. To solve the problem, an apprenticeship learning method is utilized via inverse reinforcement learning (IRL) based on both Q-learning and deep reinforcement learning (DRL). The performance of this method is compared to learning from a demonstration technique called behavioral cloning (BC) using a supervised learning approach. Simulation and numerical results show that the proposed approach can achieve expert-level performance. We also demonstrate that, unlike the BC technique, the performance of our proposed approach does not degrade in unseen situations. 

\end{abstract}


\end{frontmatter}


\section{Introduction}
\label{sec:Introduction}
The fast growth of various unmanned aerial vehicle (UAV)- use cases and applications recently attracted extensive research studies in both academia and industry. 
Different application domains utilize drones in different ways, such as cargo delivery, search and rescue (SAR), aerial imaging, disaster monitoring, surveillance and public safety, drone-assisted communication, among others \cite{andreeva2020supporting,pandey2020adaptive,javed2023uav,AfghahINFOCOM,shamsoshoara2021aerial, keshavarz2020real}. 
Unique features such as a wide ground field of view (FoV), 3-dimensional movement, fast deployment and agile response make the UAVs suitable for use in various applications. A recent report published by the Federal Aviation Administration (FAA) predicts that the number of UASs will be expected to rise from 448,000 in 2021 to 1.8 million in 2024 \cite{FAAAeros18:online,forecast2020fiscal}.

In drone-assisted communications, the UAVs appear as aerial base stations (BSs) or flying access points (APs) to serve the terrestrial User Equipment (UEs) in disaster relief scenarios. In particular, they can be used to extend the cellular coverage and improve the connectivity, or the UAVs relay the information between the UEs and the neighbor BSs \cite{mozaffari2016unmanned, mozaffari2019tutorial,Huang,zeng2016wireless,Nima_Infocom,Nima_Asilomar}.
While using drones as aerial BSs has been well studied, an emerging path focuses on cellular-connected UAVs, called \textit{aerial UEs}. In this case, the terrestrial BSs can serve the UAVs to handle both the command and control (C\&C) communication and payload communication.

Drones can be utilized as aerial UEs in LTE systems as well as 5G and beyond networks \cite{3GPP_online, 3GPP_online_subband}. However, noting the fast-growing number of unmanned aerial systems (UASs), cellular-connected drones are expected to cause various challenges to cellular systems including interference caused by UAVs to base stations and terrestrial users, spectrum management,  Quality-of-Service (QoS) management, frequent handovers, robustness, security, reliability, etc \cite{rovira2022review,9839122,lahmeri2021artificial,huang2021massive,lotfi2022}.
Moreover, traditional cellular infrastructures such as terrestrial BSs are designed to service terrestrial UEs in uplink (UL) and downlink (DL) communications and several modifications need to be applied to these systems to serve the aerial UEs \cite{huo20175g, checko2014cloud, Mohammed_ICC,9448665}. Furthermore, 
 \cite{3GPP_16,3GPP_17} recommend technical specifications to ensure reliable and efficient communication for aerial UE in 5G systems. 
For instance, path planning, beamforming, and initial cell selection are essential for enabling reliable communication in 5G-connected drones. The use of massive multiple-input multiple-output (mMIMO) technology and beamforming algorithms can improve interference management by directing the signal towards the intended receiver and minimizing interference with other users \cite{mishra2020survey}. Furthermore, an AI-based path planning model can be used at the UAVs to determine the optimal trajectory, altitude, and speed taking into account factors such as the received signal strength, coverage and level of interference. 

In this paper, we develop an interference-aware joint path planning and power allocation scheme for a cellular-connected UAV to minimize the interference it causes to the existing ground UEs in adjacent cells. The UAV is set to start its task from an initial point, and the goal is for it to reach a certain destination point to finish its task. The UAV's task can be considered as remote-sensing or aerial monitoring, where the collected information should be delivered to a fusion center using a UL communication with the desired terrestrial BS. The UAV's goal is to maximize its UL throughput and signal-to-noise ratio (SNR) while this may increase the interference on the terrestrial UEs as well. The UAV's operation field is assumed to be covered by non-overlapping cells with different UE densities, which means some cells have a higher UE  density. The UAV's flight path from the source to the destination affects both the UAV's UL throughput and the terrestrial UEs' interference. 

The problem formulation in this study accounts for several factors including the UAV's transmission power, path planning considering the terrestrial UEs density, the imposed interference to the terrestrial UEs, and the UAV's performance in terms of task completion. The performance of common reinforcement learning (RL) methods to equip the UAV with autonomous decision-making capabilities highly depends on the definition of the reward function. Defining a reward function that takes into account all the aforementioned factors is not intuitive. As a result, we proposed a novel approach based on apprenticeship learning via inverse reinforcement learning (IRL) in this paper, employing both Q-learning and deep Q-networks (DQN). We assume that an expert is available to generate some optimal datasets based on the simulator and that these datasets can be used to reconstruct the optimal reward function. The obtained reward function is then applied to find the optimal policy. To the best of our knowledge, this work is one of the leading and primary works that address the joint path planning and power allocation problem in a cellular-connected UAV using apprenticeship learning and IRL. 
The contributions of this study are as follows:
\begin{itemize}
    \item Formulating the problem of joint path planning and power allocation to minimize the UAV's UL transmission interference to the neighbor and adjacent terrestrial UEs.
    
    \item Designing an open-source simulator to collect experts' information and trajectory based on optimal behavior. The simulator is entirely object-oriented, with different classes for BSs, UEs, and UAVs. The user can change the number of cells, the density of UEs, etc. A graphical user interface is developed to show the drone's interaction with the simulation environment.
    
    \item Devising an apprenticeship learning approach using both IRL and deep IRL to analyze the optimal behavior obtained by the expert's behavior and obtain the optimal reward function. This optimal reward function can generate the optimal policy based on the user's defined threshold.
    
    \item Saving more time during the training process by using approaches such as imitation learning and apprenticeship learning which help the agent to reduce the visiting states since the optimal behavior can guide the agent on which states are necessary to visit.
    
    \item Comparing the performance of the proposed apprenticeship learning via IRL based on Q-learning and Deep Q-network with other alternative approaches such as behavioral cloning, shortest path, and random path for various scenarios including the situations where an error happens or the agent visits a state due to environmental conditions such as wind and that state has never seen before by the expert.
\end{itemize}

The rest of this paper is organized as follows. Section~\ref{subsec:related_works} reviews the related works regarding the cellular-connected UAV for similar problems and approaches. The problem definition and the system model are discussed in Section~\ref{sec:System_Model_IRL} with all related assumptions. Section~\ref{sec:method} introduces the apprenticeship learning via IRL using both Q-learning with linear function approximation and Deep Q-Network. Section~\ref{sec:Simulation} evaluates the convergence and other metrics and parameters using our designed simulator. Conclusions and discussions are summarized in Section~\ref{sec:Conclusion}.

\subsection{Related works}
\label{subsec:related_works}
This section reviews previous works regarding cellular-connected UAVs by considering different models such as interference management, power allocation, task allocation, path planning, etc. Different centralized and distributed methodologies were utilized to solve these challenges.

The authors in recent studies such as \cite{zeng2021simultaneous} considered a scenario where a cellular-connected UAV aims to finish its task and be covered by the ground cellular network; however, there are coverage gaps in the area because of antennas and infrastructure limitations such that the UAV cannot be continuously connected to the BSs. The UAV's goal was to avoid those areas during the flight to minimize the timeout probability ($P_{out}$). The authors performed the UAV trajectory/navigation optimization using a dueling double deep Q-network and RL as a model-free approach to solving this problem. The approach was considered for simultaneous navigation and radio mapping to enhance the UAV's aerial coverage while it is flying toward its destination.

In another study \cite{mei2019cellular}, the authors considered an inter-cell interference and power coordination scenario for a cellular-connected UAV. First, the authors tried to solve the problem of inter-interference minimization using a centralized approach. The authors assumed that a central point exists to collect global information for the optimization problem. The challenge is defined as a non-convex optimization problem. Hence, the authors used a successive convex approximation to solve the problem. They obtained a sub-optimal solution for this approach. Later, the authors proposed a decentralized approach by only using local information.

In a similar study~\cite{zeng2019path}, the authors defined a path planning problem for a cellular-connected UAV to minimize its mission completion time and maximize the connectivity probability to the cellular BSs. The problem is defined in a Markov decision processing (MDP) environment, and they used a temporal difference approach, as an RL, to find the optimal policy for path planning optimization. 

Most authors in this domain used a model-free approach such as the temporal difference, as an RL, to solve these kinds of interference and path planning optimization problems. However, problems like the one formulated in this paper are hard to model for dynamic environments, and even if modeled, the problem may not be convex, resulting in a suboptimal solution. Also, solving such non-convex problems needs additional resources and time to be handled. RL usually handles these complex problems. Several studies such as \cite{shamsoshoara2019distributed, shamsoshoara2019solution, shamsoshoara2020autonomous,challita2019interference} used different types of RL approaches such as distributed Q-learning, team Q-learning, multi-agents Q-learning to tackle the problem of relaying and delivering information for terrestrial UEs in disaster scenarios. 
One of the critical steps in developing a good RL solution is defining the reward function. The reward function maps a relation between the MDP environment, Q-action values, and optimal policy learning. As the problem becomes more complex, defining a meaningful reward that considers all aspects of the problem may not be trivial. In that case, the reward function is less likely to be optimal, and as a result, the obtained policy may not be optimal. Moreover, since most of the evaluated methods are implemented based on a single reward function, the comparison with other methods is less meaningful. In addition, in complex problems in a real-world scenario, it is not always possible to define a reward function that can capture all features of the problem.

With the rise of open-source software, simulators, and emulators such as network simulator-3 (NS3)~\cite{riley2010ns}, srsLTE~\cite{gomez2016srslte}, OpenAirInterface~\cite{nikaein2014openairinterface}, and other implemented simulators, it is possible to ask for experts to generate an optimal behavior for the agent in a similar simulated environment. These generated data and trajectories by the expert can be used later for the agent to learn that behavior. We like to note that in this study, the trajectory refers to the sets of states and actions in the MDP environment. The expert data can be used to define the optimal behavior, policy, or reward function. This approach is called imitation learning (IL), learning from demonstration (LfD), or apprenticeship learning (AL).
The authors of \cite{shamsoshoara2021uav} considered a disaster relief scenario where a UAV helps the terrestrial UEs to schedule a packet delivery to the neighbor BSs using an IL solution. The authors implemented a supervised learning solution as behavioral cloning (BC) approach to mimic the expert's behavior in a similar situation. BC usually does not consider any reward functions and it is only based on visited states and actions where the agent blindly follows the expert. It means that if the expert made any wrong decisions, then the agent may make the same error. Furthermore, if the visited state by the agent has never been experienced by the expert, then the agent cannot make a proper action since that state was not considered in the optimal learned policy. These challenges motivate us to utilize an approach called apprenticeship learning via IRL~\cite{abbeel2004apprenticeship}.

\subsection{Table of Notations}
\label{subsec:table_notations}
Table~\ref{tab:Notation} shows all the parameters and notations used in this study in both the system model and methodology.

\begin{table}[t!]
\caption{Notation descriptions}
 \centering{
\label{tab:Notation}
\resizebox{0.95\linewidth}{!}{  
\begin{tabular}{c|l|c|l}
\toprule
\toprule
\Topspace
\Bottomspace
\textbf{Notation} & \textbf{Description} & \textbf{Notation} & \textbf{Description}\\
\hline
\Topspace
{$\mathcal{U}$} & Set of all ground users (UEs) &{$\pmb{\varpi}$} & Path association vector
\\
{$\mathcal{U}_M$} & Set of all ground users associated to the main BSs & {$\pmb{\varrho}$} & Cell association vector
\\
{$\mathcal{U}_N$} & Set of all UEs associated to the neighbor BSs & {$\bar{T}$} & Throughput threshold
\\
{$\mathcal{B}_M$} & Set of main BSs & {$S$} & State vector in the RL approach
\\
{$\mathcal{B}_N$} & Set of neighbor BSs & {$A$} & Action vector in the RL approach
\\
{$B$} & Bandwidth & {$P_r$} & Transition probability
\\
{$u_l = (x_l, y_l, z_l)$} & UAV's longitude, latitude, and altitude & {$\gamma$} & Discount factor
\\
{$V$} & UAV's pitch speed & {$R(s)$} & Reward function
\\
{$\pmb{p}$} & UAV's taken path based on locations, i.e., $l_i$ & {$w$} & Reward function's weights
\\
{$\mathcal{L}$} & Set of all locations s.t. $l_i \in \mathcal{L}$ & {$\phi(s)$} & Features vector
\\
{$d_{a, b}$} & Distance between locations $a$ and $b$ & {$\pi(S)$} & Next action policy
\\
{$\varsigma_{b, l}$} &  SNR received at BS $b$ at location $l$ & {$V^{\pi}(s)$} & Policy value at state $s$
\\
{$f$} & Center frequency & {$\mu(s)$} & Feature expectation value
\\
{$\Phi_{b_m, l}$} & Received throughput at BS $b_m$ & {$D$} & Hyper distance
\\
{$P_{b_m, l}$} & UAV's transmission power & {$Q(s, a)$} & Q-Action value at state $s$
\\
{$h_{b_m, l}$} & Channel gain between the UAV and the BS $b_m$ & {$\theta$} & Stochastic Gradient Descent weights
\\
{$G_{b_m, l}$} & Antennas gain between the UAV and BS $b_m$ & {$MSE$} & Mean Squared Error
\\
{$N_0$} & Noise at the receiver & {$\tau$} & Tuple of states-actions trajectories
\Bottomspace
\\
\bottomrule 
\end{tabular}
} 
} 
\vspace{-10pt}
\end{table}

\section{System Model}
\label{sec:System_Model_IRL}

\begin{figure}[t!]
	\centering
	\includegraphics[width=0.8\columnwidth]{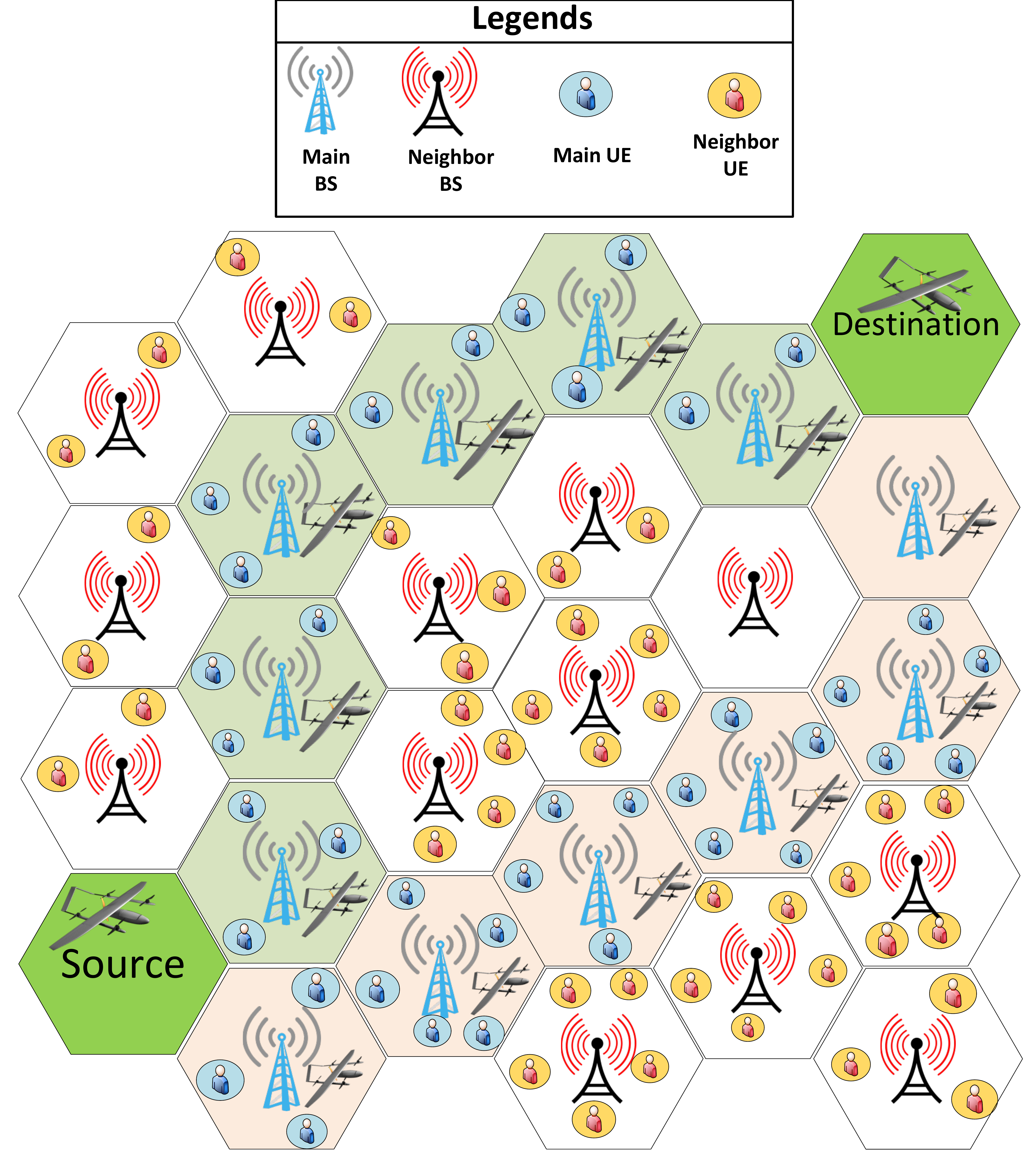}
	\caption{A sample system model showing two different possible paths of the UAV from the source to destination.}
    \label{fig:system_model}
\end{figure}
Consider a single cellular-connected UAV equipped with different payloads such as sensors, GPS, and cameras. The equipment collects data and sends it to a fusion center for further processing. We consider an orthogonal frequency-division multiple access (OFDMA) scheme for the wireless communication link, which divides the total cell bandwidth $B$ into $K$ resource blocks (RBs). The UAV utilizes an uplink (UL) to transfer its data to the ground base station (BS). Fig.~\ref{fig:system_model} depicts the system model. As seen in this figure, the UAV's operation field is divided into multiple non-overlapping hexagonal cells, each served by one BS which services several terrestrial UE with random movement. The UAV can fly over these cells to reach its destination. 
The BSs are categorized into two groups: 1) the main network to which the UAV is linked, and the BS provides the required bandwidth for the UAV's UL. UAV's transmission does not interfere with the communication of other terrestrial UEs in the same cell, 2) the neighbor BSs that the UAV is not communicating with them. However, the UAV's transmission causes interference to the neighbor BSs and their UEs when the UAV uses the same RB with them. We assumed that the neighbors' UEs cannot cause interference to the UAV's transmission since the transmission power is not strong enough. Hence, the number of terrestrial UEs in neighbor cells is critical because they are affected by the UAV's interference. Here, we consider the UAV's interference value on individual UEs in our design. 
The set $\mathcal{U}$ of U ground UEs consists of two groups of $\mathcal{U}_{\rm M} \subseteq \mathcal{U}$ and $\mathcal{U}_{\rm N} \subseteq \mathcal{U}$ which are the sets of UEs in the main and neighbor networks, accordingly. 

The base stations in the main and neighbor networks are shown as $b_{m} \in \mathcal{B}_{\rm M}$ and $b_{n} \in \mathcal{B}_{\rm N}$, respectively.
The UAV has a predefined initial source and destination. Its task is to reach the destination while keeping its communication with a legitimate base station, $b_{m}$. The UAV and other base stations are defined in a hexagon grid area where the ground BSs are fixed, and the UAV can move in six directions (North, North East, South East, South, South West, North West). While near the zone's edge, the UAV cannot fly off the grid.  
For the sake of simplicity, the BS is assumed to be in the center of each hexagon. The UAV's location is known based on the on-board GPS sensor, $u_{l} = (x_{l}, y_{l}, z_{l})$. The UAV changes its location with a constant pitch speed velocity, $V$, $0 < V \leq \bar{V}$, where $\bar{V}$ is the maximum available pitch speed for the UAV. Also, it is assumed that the UAV changes its direction using the yaw action not rolling. The UAV's path from the source to destination is shown as $\pmb{p} = (l_{\rm 1}, l_{\rm 2}, \dots, l_{\rm f})$, which the first location is the source cell, $l_{\rm 1} = S$ and the last one is the destination cell, $l_{f} = D$. All locations are subsets of the global location set $\mathcal{L}$, $(l_{\rm 1}, l_{\rm 2}, \dots, l_{\rm L}) \in \mathcal{L}$, where $L$ is the grid size or the number of hexagons. The UAV's path length is not fixed, but if it cannot reach the destination in a certain number of movements, the UAV's operation may fail due to battery limitations. Based on the literature~\cite{wu2018joint}, the grid area is assumed small enough such that the UAV's location can be assumed constant inside each hexagon. 

\subsection{Wireless Communication Model}
\label{subsec:model_com}
Buildings and obstructions have an impact on signal propagation from the air to the ground. Depending on the environment, this connection may be a line of sight (LoS) or a non-line of sight (NLoS) link. Accordingly, the path loss of LoS and NLoS links of the UAV at location $l$ and BS $b_{m}$ from the primary network on a sub-6 GHz band can be described as follows~\cite{al2014optimal}:
\begin{align}\label{eq:uav_pathloss_los}
& \varsigma_{b_m, l}^{\rm LoS}(dB) = 20\log_{\rm 10}(d_{b_m, l}) + 20\log_{\rm 10}(f) - 147.55 + \eta_{\rm LoS},
\end{align}
\begin{align}\label{eq:uav_pathloss_nlos}
& \varsigma_{b_m, l}^{\rm NLoS}(dB) = 20\log_{\rm 10}(d_{b_m, l}) + 20\log_{\rm 10}(f) - 147.55 + \eta_{\rm NLoS},
\end{align}
where $l$ is the UAV's location or the center of the cell in which the UAV is currently located. $f$ is the center frequency and $d_{b_m, l}$ is the distance between BS $b_m$ and the UAV located at location $l$. Also, $\eta_{\rm LoS}$ and $\eta_{\rm NLoS}$ are LoS and NLoS links attenuation factors, respectively. 
The probability of having LoS link depends on environmental factors. In our model, for UAV to the ground link, the LoS probability is given by the proposed model in~\cite{al2014optimal}:
\begin{align}\label{eq:los_probability}
\mathcal{P}_{\rm LoS} (d_{b_m,l})=\frac{1}{1+c_1 \exp{\bigg(-c_2\bigg(\frac{180}{\pi}\tan^{-1}(\frac{H}{d_{b_m,l}})-c_1\bigg)\bigg)}},
\end{align}
where $H$ represents the UAV height, and $c_1$ and $c_2$ are environmental dependent permanents.
Accordingly, the total path loss of UAV at location $l$ and ground BS $b_m$ can be defined as 
\begin{align}\label{eq:total_pathloss}
\varsigma_{b_m, l} = \mathcal{P}_{\rm LoS}\,\,\varsigma_{b_m, l}^{\rm LoS} + \big(1-\mathcal{P}_{\rm LoS}\big) \,\,\varsigma_{b_m, l}^{\rm NLoS}.
\end{align}

The SNR for the UAV UL data transmission to the main BS $b_m$ is written as $\Phi_{b_m, l}$: 
\begin{align}\label{eq:uav_sinr}
& \Phi_{b_m, l,k} = \frac{P_{b_m, l,k}h_{b_m, l,k}}{N_0},
\end{align}
where $P_{b_m, l, k}$ is the transmission power that the UAV allocates to each RB and sends its data to BS $b_m$ at location $l$. $h_{b_m, l,k} = G_{b_m, l,k} 10^{-\varsigma_{b_m, l}/10}$ indicates the channel gain between the UAV and the main BS $b_{m}$ based on the path loss defined in \ref{eq:total_pathloss}, where $G_{b_m, l,k}$ denotes the antenna gain values of the UAV and the BS ($b_{m}$). While we assumed a simplified model in this study to calculate the gain value, in the real-world scenario, the antennas of the BSs are down-tilted to have better coverage for the ground UEs and reduce the effect of inter-cell interference. Hence, the UAV is usually served by the side-lobe antenna of the ground BS. $N_{\rm 0}$ is the noise power spectral density at the main receiver BS $b_{m}$. 
Based on the SNR definition in (\ref{eq:uav_sinr}), and by considering $B_k$ as the bandwidth of RB $k$, the throughput rate for the UAV at location $l$ and based on the connected BS $b_{m}$ can be written as:
\begin{align}\label{eq:uav_rate}
& T_{b_m, l} =\sum_{k=0}^{K} B_k \log_{\rm 2}(1 + \Phi_{b_m, l,k}).
\end{align}

The UAV may interfere with the terrestrial UEs transmission in the neighbor networks. 
This interference depends on the UAV's transmission power, 
quality of UAV's channel to the neighbor BS $b_n$
and the probability of allocating the same resource block in the UAV and terrestrial UEs. Here, due to minimum knowledge sharing the UAV is blind to the knowledge of UEs in the neighbor networks. Therefore, we assume that there is at least one UE in neighbor cells that is assigned with the  same resource block. Accordingly, the UAV's interference in uplink of a UE $u \in \mathcal{U}_{\rm N} \subseteq \mathcal{U}$ at BS $b_{n}$, and by considering $\mathcal{K}_u$ as a set of allocated RBs to UE $u$,   
will be~\cite{challita2019interference}:
\begin{align}\label{eq:ue_interference}
    I_{b_n,l}^{u} = \sum\limits_{m=1}^{\mathcal{B}_{\rm M}} \sum\limits_{k=0}^{|\mathcal{K}_u|} P_{b_m,l,k}h_{b_n, l,k},
\end{align}
where $P_{b_m, l,k}$ is the UAV's transmission power in RB $k\in \mathcal{K}_u$ when it was transmitting to the main BS $b_{m} \in \mathcal{B}_{\rm M}$ at location $l$ on the same RBs of UE $u$. Also, $h_{b_n, l,k}$ indicates the channel gain between the UAV and the neighbor BS $b_n$. This interference rate is an argument that affects the terrestrial UE $u$ SINR and throughput rate: 
\begin{align}\label{eq:ue_rate_sinr}
& \Phi_{b_n}^u = \frac{{P_{b_n}^u h_{b_n}^u}}{I_{b_n,l}^u + N_{\rm 0}}, 
\\
& T_{b_n}^u = B \log_{2}(1 + \Phi_{b_n}^u),
\end{align}
where 
{$P_{b_n}^u$} is the transmission power for UE $u$, for the sake of simplicity, all UEs' transmission powers are assumed to be constant. Also, $h_{b_n}^u$ indicates the channel gain between the UE $u$ and the BS $b_n$.  

\subsection{Objective Definition}
\label{subsec:model_obj}
Based on the defined system model, the UAV's objective is to fly from the source point and reach the destination point based on the BSs' locations and the density of the UEs in neighbor cells to minimize the interference level it causes on the neighbor terrestrial UEs, maximize its UL throughput by finding an optimal transmission power value, and the shortest path between the source and destination. 

The path association vector based on the hexagons is shown as $\pmb{\varpi}$ that consists of elements $\varpi_{a, b} \in \{0, 1\}$. The value 1 means that the UAV has moved from location $a$ to $b$ and 0 means otherwise. The UAV has to find the optimized path association vector from the source to destination $\pmb{\varpi^{*}}$.
To meet these requirements, the UAV has to find the optimal transmission power $P_{b_m, l}^{*}$ at each location $l$ where the UAV is associated with tower $b_m$ which is the main base station. The optimal transmission power $P_{b_m, l}^{*}$ for transmission to base station $b_m$ can be chosen from the transmission power range $P_{b_m, l}^{*} \in [P_{\rm min}, P_{\rm max}]$ which includes a vector of discrete values. $P_{b_m, l}^{*}$ should meet the satisfactory throughput rate $T_{b_m, l} > \bar{T}$, where $\bar{T}$ is the minimum acceptable throughput. 
The optimal power vector during the UAV's flight is shown as $\pmb{P^{*}}$.
To find $P_{b_m, l}^{*}$, the UAV has to find its BS-connectivity association vector. This connectivity vector, $\varrho_{b_m,l} \in \pmb{\varrho}$, shows  whether the UAV connects to BS $b_m$ at location $l$ or not. The values for the connectivity vector are $\{\rm 0, 1\}$.  

The UAV has to find the best BS connectivity vector $\pmb{\varrho^{*}}$ based on other optimization variables in this problem. Since the UAV does not cause any interference on the UEs in its current cell and it can interfere with other neighbor UEs in neighbor cells, the optimal variables such as $\pmb{\varpi^{*}}$, $\pmb{P^{*}}$, and $\pmb{\varrho^{*}}$ are chosen to minimize the interference to neighboring terrestrial UEs. 
In a nutshell, this optimization problem can be written as:
\textcolor{blue}{}
\begin{subequations}
\begin{align}
\nonumber
\min_{\pmb{P}, \pmb{\varpi}, \pmb{\varrho}}  \hspace{0.5cm}&\alpha \, 
\sum_{l = l_1}^{l_L} \sum_{n=1}^{|\mathcal{B}_{\rm N}|} I_{b_n,l}^u+ \beta \sum_{a=l_1}^{l_L}\sum_{b=l_1}^{l_L}\varpi_{a, b},\\
&+ \delta \sum_{l=l_1}^{l_L}\sum_{m=1}^{|\mathcal{B}_{\rm M}|}\varrho_{b_m, l} , \label{eq:obj_main}
\\
\text{s.t.,} 
& \hspace{0.3cm}P_{\rm min} \leq P_{b_m, l} \leq P_{\rm max},\,\,b_m \in \mathcal{B}_{\rm M}, 
\label{eq:obj_8}
\\
& \hspace{0.3cm} \varpi_{a, b} \in \{0, 1\}, \,\, \varrho_{b_m,l} \in \{0, 1\}, \,\, l,a, b \in \mathcal{L}, \label{eq:obj_7}
\end{align}
\begin{align}
&\hspace{0.3cm} \sum_{l=l_1}^{l_L}T_{b_m, l} \geq \varrho_{b_m, l} \bar{T}, 
\label{eq:obj_1}
\\
& \hspace{0.3cm}\sum_{m=1}^{|\mathcal{B}_{\rm M}|}\varrho_{b_m, l} - \sum_{b=l_1, b\neq a}^{l_L}\varpi_{b, a} = 0, 
\label{eq:obj_2}
\\
& \hspace{0.3cm}\sum_{a=l_1}^{l_L}\varpi_{S, a} = 1, \quad \quad 
a \neq S,  \label{eq:obj_3}
\\
& \hspace{0.3cm} \sum_{b=l_1}^{l_L}\varpi_{b, D} = 1, \quad \quad 
b \neq D, \quad b \neq S, \label{eq:obj_4}
\\
& \hspace{0.3cm} \sum_{a=l_1, a\neq b}^{l_L}\varpi_{a, b} - \sum_{k=l_1, k\neq b}^{l_L}\varpi_{b, k} = 0, \label{eq:obj_5}
\\
& \hspace{0.3cm}\sum_{a=l_1, a\neq b}^{l_L}\varpi_{a, b} \leq 1. 
\label{eq:obj_6}
\end{align}
\end{subequations}
The optimization problem in (\ref{eq:obj_main}) aims to minimize the interference level on neighbor network UEs and BSs, also it minimizes the UAV's path length from the source to the destination and selects the main BSs for the communication which imposes less interference to neighbor networks.
Equations (\ref{eq:obj_8}) and (\ref{eq:obj_7}) are the optimization problem constraints regarding the UAV's transmission power, the UAV's path between different hexagon cells, and the UAV-BS connectivity vector. Equation (\ref{eq:obj_1}) guarantees that the throughput rate for the UAV-BS UL at each location $l$ when it is associated with BS $b_m$ is greater than a predefined threshold $\bar{T}$, equation (\ref{eq:obj_2}) guarantees that the UAV is only connected to one main BS $b_m$ at each location $l$ on its path to the destination. Equation (\ref{eq:obj_3}) claims that the UAV's path starts with the source point $S$ and the UAV only visits the source one time, (\ref{eq:obj_4}) claims that each UAV's path is ended at the destination $D$, (\ref{eq:obj_5}) guarantees that if the UAV visits any midpoint such as $b$ between the source $S$ and destination $D$, then it will leave that point and fly to another cell. (\ref{eq:obj_6}) claims that each hexagon $b$ on the UAV's path is going to be visited at most one time. 

Fig.~\ref{fig:system_model} shows a sample scenario of this system model. The UAV chooses two paths from the source to the destination, the green one is the optimal path considering the density of the terrestrial UEs since the neighbor cells have lower densities and the UAV's UL communication has less interference on the terrestrial UEs. We should note that the UAV can allocate an optimal value for the transmission power in each cell to control the trade-off between the throughput satisfaction threshold and the level of interference. On the other hand, the orange path shows a route in which the neighbor cells carry a larger density regarding the number of UEs, and as a result, the UAV faces a more strict restriction on its transmission power. 

This optimization problem is categorized as a mixed integer non-linear one which is hard to solve. This study aims to consider an autonomous approach as a solution for the UAV to find the optimal values for the UAV's transmission power, path, and cellular connectivity vector. This multi-parameter problem
involves a huge state-space set which is overwhelming to solve and find the optimal values in a reasonable amount of time. To investigate the possible solution for this problem, we utilize an apprenticeship learning approach and use the expert's knowledge to expedite the solving problem and have a better understanding of the optimal reward function.

\section{UAV's Self-Organizing Approach}
\label{sec:method}
\begin{figure*}
    \centering
    \begin{subfigure}{0.48\linewidth}
        \centering
        \includegraphics[width=0.8\textwidth]{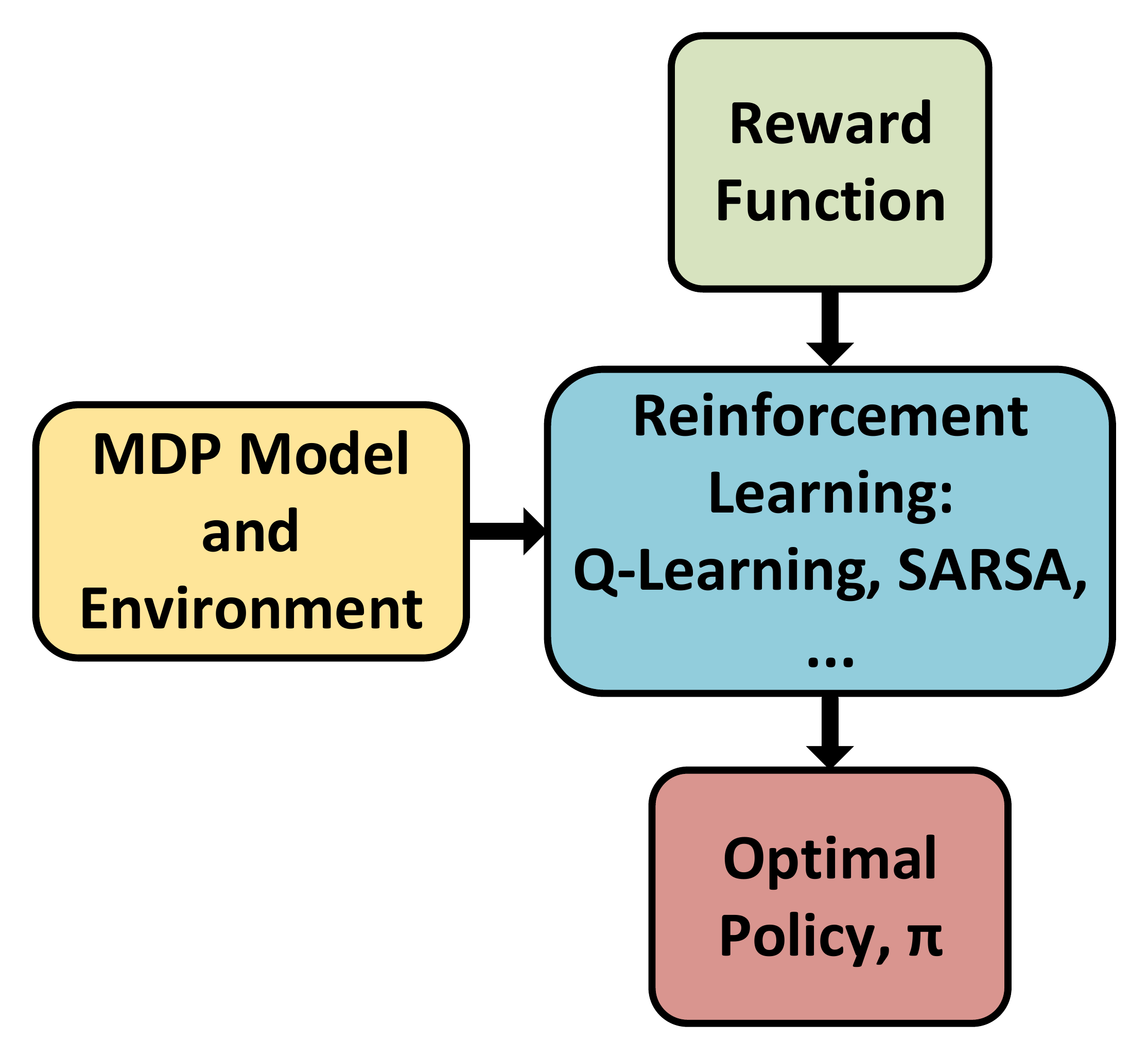}
        \caption{Reinforcement learning approach}
        \label{subfig:AL_RL}
    \end{subfigure}
    \hfill
    \begin{subfigure}{0.48\linewidth}
        \centering
        \includegraphics[width=\textwidth]{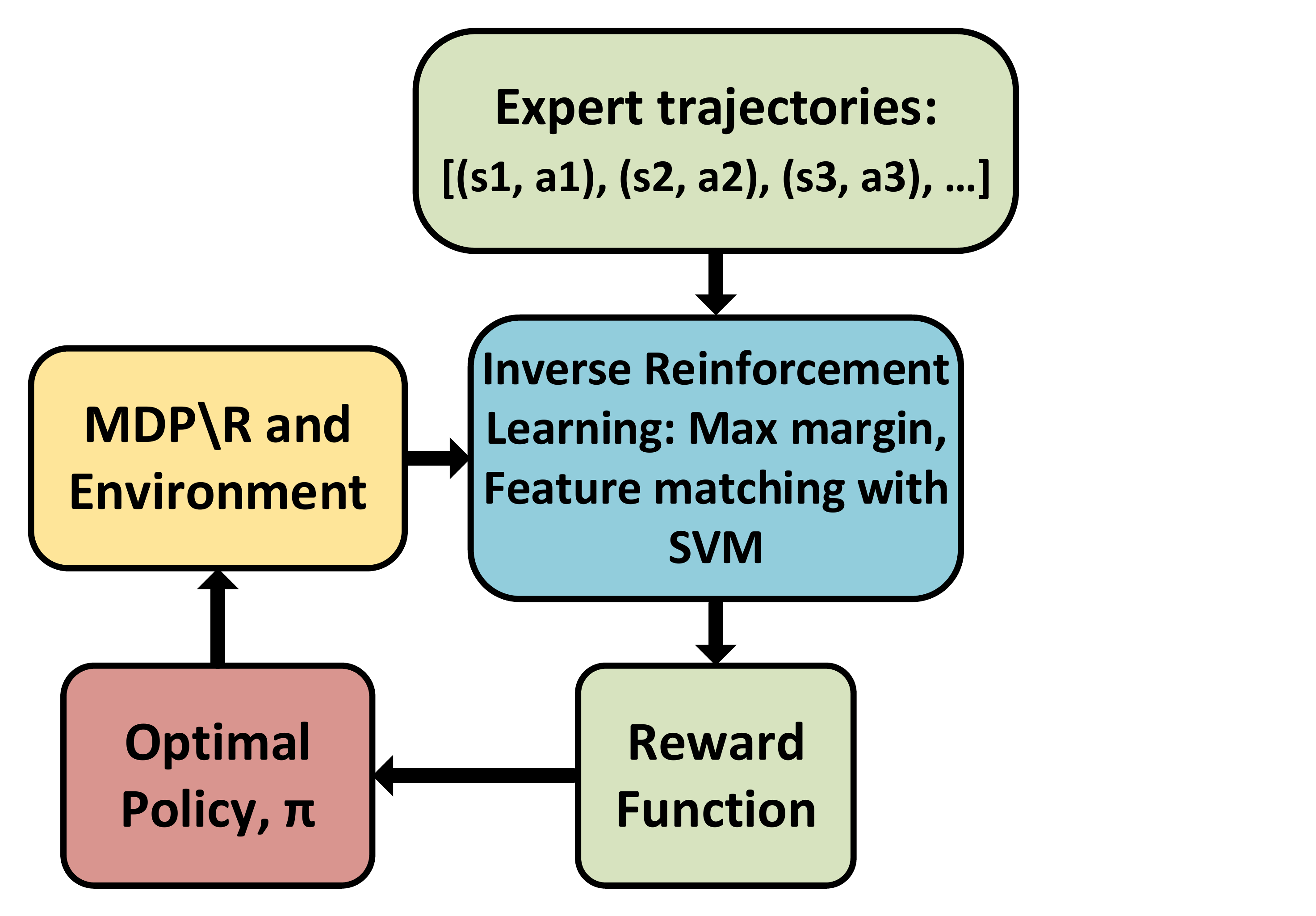}
        \caption{Apprenticeship learning using Inverse RL}
        \label{subfig:AL_IRL}
    \end{subfigure}
    \caption{Difference between reinforcement learning and Inverse RL for apprenticeship learning (SARSA: State–action–reward–state–action, SVM: support vector machine).}
    \label{fig:RL_VS_IRL}
\end{figure*}

Many artificial intelligence and robotic applications consist of autonomous agents such as ground robots, UAVs, and self-driving cars. These agents usually operate in an uncertain and complex environment in which the decision-making task is cumbersome and necessary.

In multi-objective reinforcement learning problems, where the agent deals with various and possibly conflicting objective functions to perform tasks of different natures, it is not always possible to define a reward function that optimally captures all the objectives ((\ref{eq:obj_main}) - (\ref{eq:obj_6})). The problem studied in this paper also involves multiple actions such as the movement and level of transmission power with different natures and behavior. This problem makes it more intense for the agent to have a meaningful understanding of the reward function.
Usually, in these situations, an expert who has enough experience in the domain and field can demonstrate appropriate behavior to the agent. These demonstrations may save some training time while also assisting the agent in determining the best reward function based on the most optimal presented behavior. This type of learning is referred to as Learning from Demonstration (LfD) or Programming by Demonstration (PbD), and it is also known as imitation learning (IL).
IL techniques are categorized into two branches, i) behavioral cloning (BC) \cite{bain1995framework} and ii) apprenticeship learning via inverse reinforcement learning (IRL) \cite{osa2018algorithmic,adams2022survey}.

In BC, the agent blindly mimics the expert or teacher without considering the expert's optimality which means if the expert makes a wrong decision, the expert may follow that. Usually, IL problems are solved using supervised learning approaches. However, if the observed states by the agent have not been experienced by the expert, the learned policy may not achieve an optimal outcome. This issue can be solved by an approach called Data Aggregation (DAGGER) \cite{ross2011reduction}. DAGGER aims to collect expert demonstrations based on the state distribution observed by the agent. In this case, the expert should be available at all times, which is not possible in all scenarios and applications. 

On the other hand, apprenticeship learning via IRL tries to find the expert's hidden desire (reward function) from the demonstrations to find the optimal policy \cite{ng2000algorithms, abbeel2004apprenticeship}. This paper utilizes apprenticeship learning via IRL to solve the problem proposed in Section~\ref{sec:System_Model_IRL}. Similar approaches such as BC, Q-Learning, and Deep Q-Network are also implemented to evaluate the numerical performance.  

\subsection{Apprenticeship Learning via Inverse Reinforcement Learning}
\label{subsec:IRL}

The problem definition of this study can be shown as a sequential decision-making process that can be posed in the (finite-state) Markov decision process (MDP). Typically, the reward function is provided in MDP problems, allowing the agent to determine the optimal transition probabilities, value function, and optimal policy. 
The agent can utilize the optimal policy for decision-making in every state. However, the assumption that the reward function is present in every complex system model is overly optimistic, and manually obtaining the reward function is challenging.

The existence of an expert may make it possible to have a set of trajectories and demonstrations to learn the optimal behavior. The expert could be a virtual agent that operates on a high-performance computing machine but cannot be utilized onboard a UAV. In this study, the term ``\textit{trajectory}'' refers to a set of states and actions rather than the UAV's trajectory movement or path. The IRL aims to re-construct the unknown reward function $R(\tau)$ from the expert demonstrations. Later, this obtained reward function can be used in an RL approach to find the optimal policy. But, this problem is ``\textit{ill-posed}'' since there exist multiple reward functions for a unique and optimal policy. 

Here, we assumed that the expert is trying to optimize an unknown reward function which this unknown function can be represented as a linear mixture of known ``\textit{features}''. 
Fig.~\ref{fig:RL_VS_IRL} shows the main difference between a reinforcement learning problem and an apprenticeship learning via IRL (AL-IRL) problem. The reward function is available in Fig.~\ref{subfig:AL_RL}, and based on that the optimal policy is gained. However, in Fig.~\ref{subfig:AL_IRL}, the reward function is found based on the IRL and the expert data, and then the optimal policy is obtained. Expert trajectories are not the same as optimal policies; however, they describe optimal behavior.

We assumed that the problem is MDP\textbackslash R which shows the MDP problem without a reward function. This MDP\textbackslash R consists of a tuple of $\langle \mathcal{S}, \mathcal{A}, Pr, \gamma, T \rangle$, where $\mathcal{S}$, $\mathcal{A}$, and $Pr$ represent the state space, action space, and transition probability from current state $s\in \mathcal{S}$ to the next state $s' \in \mathcal{S}$, respectively. Also, $\gamma$ is a discount factor, and $T$ is a set of trajectories demonstrated by an expert. The MDP\textbackslash R tuples are described as follows:\\
\noindent{1) State}

To define the state $s\in\mathcal{S}$, we assume that a vector of features $\boldsymbol{\phi}$ : s $\rightarrow$ [0, 1]$^k$, where $k$ is the number of features. $\mathcal{S}$ represents a finite set of feature states, $\mathcal{S}=\{\phi_i | 1\leq i \leq 5 \}$. Here, each state $s \in \mathcal{S}$ is displayed with five feature functions.
\begin{itemize}
    \item $\phi_1(s)$ is the feature that keeps track of the UAV's distance to the destination. This feature helps the agent to have a better understanding of the shortest path.
    \item $\phi_2(s)$ is the feature that keeps track of the UAV's hop counts as its traveled distance. This feature helps the agent to perform its task in a limited number of hop counts because of UAV’s battery life constraint.
    \item $\phi_3(s)$ keeps track of the UAV's successful task, where the term "successful" refers to the cases in which the UAV reaches the destination cell. $\phi_3(s)$ behaves like a binary variable, where it takes the value of "0" if the UAV doesn't reach the destination and "1" for reaching the destination cell. 
    \item $\phi_4(s)$ is a feature for the amount of received throughput from the UAV to the designated BS based on the allocated transmission power and SNR defined in (\ref{eq:uav_sinr}) and (\ref{eq:uav_rate}) at each state.
    \item $\phi_5(s)$ keeps the value for the amount of interference imposed on the neighbor UEs located in the neighbor cells.
\end{itemize}
All feature values are normalized to be in the range of [0, 1]. This condition on the reward function is necessary for the IRL algorithm to converge. Hence, each state can be expressed based on the feature value functions as, 
\begin{align}\label{eq:uav_feature_value}
\boldsymbol{\phi}(s_t) = \{\phi_i(s_t), 1\leq i \leq 5 \}.
\end{align}

\noindent{2) Action}

$\mathcal{A}$ represents the action space which consists of the UAV's movement action and the transmission power allocation $\pmb{P}$. The UAV's movement action somehow includes the defined movement variables $\pmb{\varpi}, \pmb{\varrho}$ in \eqref{eq:obj_main}. The UAV chooses both actions autonomously at the same time. \\
\noindent{3) Reward}

Here, the unknown reward function is defined by considering the feature functions in the state space. The reward function is defined as:
\begin{align}\label{eq:uav_reward_irl}
& R(s) = \boldsymbol{w}^* \ . \ \boldsymbol{\phi}(s),
\end{align}
where, $\boldsymbol{w}^*$ are the optimal weights for the features to define the desirable optimal reward function, $\boldsymbol{w} \in \mathbb{R}^k$. To have the convergence for the IRL, it is necessary to bound the reward function by $-1$ and $1$. Therefore, the weights are bounded as well $||\boldsymbol{w}^*|| \leq 1$. 
Based on the definition in (\ref{eq:uav_reward_irl}) and the defined features, the reward function can be re-written as: 
\begin{align}\label{eq:uav_reward_irl_2}
R(s, a, s') & = \boldsymbol{w}^T \ . \ \boldsymbol{\phi}(s_t),
\\ \nonumber
& = \sum_{i=1}^{5}w_i \phi_i(s, a, s').
\end{align}

As described in  (\ref{eq:uav_reward_irl_2}), the reward is a linear combination of the feature values, $\boldsymbol{\phi}(s_t)$.
It is worth mentioning that not all feature value functions behave positively for the reward function. For instance, the throughput acts as an encouragement, and the interference plays a punishment role. The goal of the IRL algorithm is to find the optimum value for the weight vector, $\boldsymbol{w}^*$, to express the expert's behavior for this specific problem. Now, let's define $\pi$ as a policy that maps the visited states to probabilities on the action vectors for the decision-making process. The value of each policy $\pi$ can be defined as an expected value of the summation of discounted reward values based on the chosen policy:
\begin{align}\label{eq:uav_value_function}
\mathbb{E} \ [V^{\pi}(s)] &= \mathbb{E}_{\pi}\ \left[\sum_{i=0}^{\infty} \gamma^{i} R (s_i, a_i, s'_i) \right] ,
\nonumber
\\
&=  \mathbb{E}_{\pi}\ \left[\sum_{i=0}^{\infty} \gamma^{i} \ \boldsymbol{w} \ . \ \boldsymbol{\phi}(s_i, a_i, s'_i) \right] ,
\nonumber 
\end{align}
\begin{align}
&= \boldsymbol{w} \ \mathbb{E}_{\pi}\ \left[\sum_{i=0}^{\infty} \gamma^{i}\boldsymbol{\phi}(s_i, a_i, s'_i) \right],
\\ \nonumber
&= \boldsymbol{w} \ \mu(\pi), 
\end{align}
where $i$ is the step that the agent or the UAV takes in every episode. $\mu(\pi)$ is defined as a feature expectation value for the policy $\pi$, $\mu(\pi) \in \mathbb{R}^k$, and $k$ is the number of defined features. 

To find the unknown parameters, $\boldsymbol{w}$, in the IRL algorithm, different tools such as feature expectation matching~\cite{abbeel2004apprenticeship, osa2018algorithmic}, maximum margin planning~\cite{ratliff2006maximum, wulfmeier2015maximum}, and maximum causal entropy~\cite{zhou2017infinite} have been proposed. In this paper, we use the feature expectation matching between the expert and the learner UAV. The reason is that the behavior of the agent and expert can be expressed based on different explicit features, hence feature expectation matching can be used in this scenario.

We assume that there exists an expert having access to the simulator environment. The expert has a full understanding and knowledge of the problem and scenario which means it considers the UEs' density, shortest path, throughput, and interference for the sake of the optimization problem defined in (\ref{eq:obj_main}) - (\ref{eq:obj_6}). The expert simulates a finite number of trajectories to show the desired behavior. These expert behaviors can be saved in terms of trajectories as some vector of features state $\boldsymbol{\phi}(s)$ and actions $a(s)\in \mathcal{A}$. Later, these trajectories can be used to define the expert feature expectation value by getting an average over several trajectories as follows:
\begin{align}\label{eq:expert_feature_expectation}
& \bar{\mu}_{E} = \frac{1}{\rm N} \sum_{j = 1}^{\rm N} \ \sum_{t=0}^{\infty} \gamma^{t} \phi(s_t^j),
\end{align}
where $N$ is the number of trajectories and $j$ denotes the index for each trajectory. This $\mu_E$ or $\mu(\pi_E)$ carries the expert policy for the demonstrated trajectories.

\textbf{Claim:} The goal is to find a policy $\pi^{*}$ such that the second-norm distance between the expert feature expectation and the current policy feature expectation is less than a threshold $\epsilon_{\textnormal{IRL}}$. The mentioned policy will meet the criteria for the value of that policy as well which means the first-norm distance between the two value functions for the expert policy and learned policy based on the unknown reward function is less than the same threshold $\epsilon_{\textnormal{IRL}}$.

\begin{proof}
The first-norm distance between the expert's and agent's value function is written as, 
\begin{align}\label{eq:value_function_distance}
\textnormal{D} &= \abs{\mathbb{E} \ \left[V^{\pi_{\rm E}}\right] - \mathbb{E} \ \left[V^{\pi^*}\right]},\nonumber
\\ 
&= \abs{\mathbb{E}_{\pi_{\rm E}}\left[\sum_{i=0}^{\infty} \gamma^{i} R (s_i, a_i, s'_i)   \right] - \mathbb{E}_{\pi^*} \left[\sum_{i=0}^{\infty} \gamma^{i} R (s_i, a_i, s'_i)  \right]},\nonumber
\\ 
&= \abs{\boldsymbol{w}^{T}\mu(\pi^*) - \boldsymbol{w}^{T}\bar\mu(\pi_{\rm E})},\nonumber
\\ 
&\leq \norm{\boldsymbol{w}}_2 \norm{\mu(\pi^*) - \bar\mu(\pi_{\rm E})}_2,\nonumber
\\
& \leq 1 \ \ . \ \ \epsilon_{\textnormal{IRL}},
\\ \nonumber
&\leq \ \epsilon_{\textnormal{IRL}}.
\end{align}
This shows that finding the optimal weights for the reward function such that $||\mu(\pi^*) - \bar\mu(\pi_E)||_2 \leq \epsilon_{\textnormal{IRL}}$ guarantees that the distance between the value function for the expert and the optimal policy is less than the same threshold. 
\end{proof}

\subsection{{Support Vector Machine Problem Formulation}}

To find the weights for the reward function, the authors of \cite{abbeel2004apprenticeship} defined the problem as a support vector machine (SVM) problem, where the aim is to maximize the difference between the value function for the expert optimal policy and other previous learned policies. In the SVM problem, this maximization problem is mapped to maximize the distance between hyperplanes. The goal of this SVM problem is to find a hyper-distance or margin ($D$) such that the difference between the expert value function and all other previously learned policies is $D$. This SVM problem can be written as
\begin{align}\label{eq:SVM_problem}
& \abs{\boldsymbol{w}^{*^T}\mu(\pi_i) - \boldsymbol{w}^{*^T}\bar\mu(\pi_E)} \leq \Delta,
\\
& \boldsymbol{w}^{*^T}\abs{\mu(\pi_i) - \bar\mu(\pi_E)} \leq \Delta,
\end{align}
where, $i$ is index of the $i^{th}$ learned policy using the reinforcement learning approach. To solve this SVM problem, the class for the expert policy is labeled as ``$+1$'' and all other learned policies from the RL tool are labeled as ``$-1$''. Based on the defined SVM problem, the distance between the origin and the expert policy class is $\frac{1}{\norm{\boldsymbol{w}}_2}$ and the distance between the origin and all other learned policies is $\frac{-1}{\norm{\boldsymbol{w}}_2}$, hence, maximizing the distance between hyper-planes is equal to minimizing $\norm{\boldsymbol{w}}^2_2$. As a result, (\ref{eq:SVM_problem}) can be written as:
\begin{subequations}
\begin{align}
& \min_{\boldsymbol{w}} \qquad \norm{\boldsymbol{w}}_2^2 \ ,
\label{eq:SVM_problem_QP_OBJ}
\\ 
& \ s.t.  \qquad \ \boldsymbol{w}^T \mu_E \geq 1 ,
\label{eq:SVM_problem_QP_CONS1}
\\
& \qquad \qquad \boldsymbol{w}^T \mu_{\pi_i} \leq -1~.
\label{eq:SVM_problem_QP_CONS2}
\end{align}
\end{subequations}


Since this SVM problem looks like a quadratic programming (QP) problem, any convex optimization tools or QP solvers such as CVXPY~\cite{diamond2016cvxpy} or CVXOPT~\cite{Home_CVX87:online} can solve this optimization problem. Here, CVXOPT is used in Python to find the weights. The objective function of this optimization, (\ref{eq:SVM_problem_QP_OBJ}), is to minimize the weights and the variables are the weights as well. Constraint (\ref{eq:SVM_problem_QP_CONS1}) is the subject for the expert policy with the label of ``+1'' and constraint (\ref{eq:SVM_problem_QP_CONS2}) is the subject for all the learned policies at different iterations with the label of ``-1''. It is worth mentioning that the algorithm of finding the optimal weight such that the distance is less than a threshold $\epsilon_{\textnormal{IRL}}$ is iterative and the number of learned policies is increasing in the QP, hence the optimality of the given solution is crucial in this problem. The standard format of any QP problem looks like the following,
\begin{subequations}
\begin{align}
& \min_{x} \qquad \frac{1}{2} x^T P x + q^T x ,
\label{eq:problem_QP_OBJ}
\\ 
& s.t. \qquad \quad Gx \leq h ,
\label{eq:problem_QP_CONS1}
\\
& \qquad \qquad \ Ax = b~.
\label{eq:problem_QP_CONS2}
\end{align}
\end{subequations}
To match this with our SVM problem (\ref{eq:SVM_problem_QP_OBJ}), matrices of $q$, $A$, and $b$ are all zeros and matrix $P$ is $2\mathbf{I}_K$, which $\mathbf{I}$ is the identity matrix and $K$ is the number of features (K=5 in our problem). The dimension of matrices $G$ and $h$ changes and increases with each iteration. 
Then in $n^{th}$ iteration for different weights and learned policies, the formats of $G$ and $h$ are
\begin{flalign}\label{eq:matrix_G}
G_{(n+1, k)} = &
\begin{bmatrix}[cccc]
   -\mu_E(1) & -\mu_E(2) & \hdots & -\mu_E(k)
   \\
   \mu_{\pi_1}(1) & \mu_{\pi_1}(2) & \hdots & \mu_{\pi_1}(k) \\ 
   \vdots & \vdots & \vdots \\ 
   \mu_{\pi_n}(1) & \mu_{\pi_n}(2) & \hdots & \mu_{\pi_n}(k)
\end{bmatrix},
\end{flalign}

\begin{flalign}\label{eq:matrix_h}
h_{(n+1, 1)} = &
\begin{bmatrix}[c]
   -1 \\
   -1 \\ 
   \vdots \\ 
   -1
\end{bmatrix},
\end{flalign}
where $\mu_{\pi_n}(k)$ is the $k^{th}$ UAV's feature expectation based on the $n^{th}$ learned policy. These matrices of $P$, $q$, $G$, $h$, $A$, and $b$ map the QP problem to our scenario defined in (\ref{eq:SVM_problem_QP_OBJ}), (\ref{eq:SVM_problem_QP_CONS1}), and (\ref{eq:SVM_problem_QP_CONS2}).
Algorithm \ref{algo:AL_IRL} summarizes the approach of the apprenticeship learning via IRL inside the loop iteration to find the optimal weights. 
\begin{algorithm}
\SetAlgoLined
 \textbf{Initialization:}\\
 Load the simulation environment\\
 Load the expert trajectories\\
 Calculate the expert feature expectation
 $\bar\mu_{E} = \frac{1}{N} \sum\limits_{j = 1}^{N} \ \sum\limits_{t=0}^{\infty} \gamma^{t} \phi(s_t^j)$\;
 Add some random values to the agent feature expectation for the first round of optimization\\
 \While{True}{
  Weights = SolveQP(Expert feature exp, Agent feature exp list)\;
  Weights($w_i$) = Weights / norm\;
  TrainedPolicy($\pi_i$) = \textbf{learner}(Weights)\;
  Reset the simulation environment\;
  $\mu_i(\pi_i, w_i)$ = RunSimulation($\pi_i$)\;
  Add $\mu_i(\pi_i, w_i)$ to the agent feature expectation list\;
  Hyper distance = D  = $\abs{w^{T}\mu(\pi_i) - w^{T}\bar\mu(\pi_E)}$\;
  \If{D $<$ $\epsilon_{\textnormal{IRL}}$}{
  Break\;
  }
 $i += 1$\;
 }
 \textbf{Return} Weight($w_i$), $\mu_i$, $\pi_i$\;
 \caption{Apprenticeship learning via IRL algorithm}
 \label{algo:AL_IRL}
\end{algorithm}
The main loop of Algorithm~\ref{algo:AL_IRL} continues until it reaches a point where the hyper distance is less than a threshold and in that case, it breaks from the main loop and returns the optimal weights $(w_i)$, feature expectation ($\mu_i$), and the learned policy ($\pi_i$). 

To train a model for the optimized weights of the reward function and find the optimal policy to evaluate that for the hyper distance, it is required to utilize learning methods like reinforcement learning (RL), especially a temporal difference (TD) learning tool. These methods find the optimal policy on line 9 of the Algorithm~\ref{algo:AL_IRL}. In this paper, we use two TD methods, the first one is Q-learning based on linear function approximation and the second one is deep reinforcement learning (DRL) or deep Q-network (DQN). The next two sections propose these two tools for the mentioned problem of apprenticeship learning to learn the policy. 

\subsection{AL-IRL with Q-learning using Linear Function Approximation}
\label{subsubsec:IRL_QL_LFA}
After finding the normalized weights in apprenticeship learning, the reward function is known for the specific iteration and the problem of MDP\textbackslash R is converted to a MDP problem with the tuple of $\langle \mathcal{S}, \mathcal{A}, Pr, R, \gamma\rangle$ where $R$ is the reward function defined based on the obtained weights, (\ref{eq:uav_reward_irl}) and (\ref{eq:uav_reward_irl_2}). The learning policy aims to find a policy that can decide on an action for the agent at each observed state to maximize the immediate reward with also considering the future reward. Different value iteration approaches are available to find this policy~\cite{sutton2018reinforcement, levine2020offline}. We chose the Q-learning tool to solve this problem \cite{shoham2003multi, busoniu2006multi}. In the standard Q-learning, the updating rule for the Q values is defined based on
\begin{align}
\label{eq:Q_learning_update_discrete}
& Q(s, a) \leftarrow (1-\alpha)Q(s, a) + \alpha[R(s, a, s') + \gamma \max_{a'}Q(s', a')],
\end{align}
where $s\in\mathcal{S}$ is the current state, $a\in \mathcal{A}$ is the action chosen based on the policy or random exploration which changes the agent's state from $s$ to $s'$, $\alpha$ is the learning rate, $R(s, a, s')$ is the reward function defined based on the IRL approach from the previous section. $\gamma$ is the discount factor considering the future reward (value-iteration or $Q(s')$). After updating the Q-values based on different approaches such as temporal difference (TD) or Monte Carlo (MC), the agent follows the policy rule for the exploitation greedy behavior. Later, this policy is considered as a learned policy for the apprenticeship learning.
\begin{align}
\label{eq:Q_learning_policy}
& \pi(s) = \argmax_a Q(s, a) ~.
\end{align}

It is worth mentioning that this Q updating approach works for problems with discrete states and actions such as the grid-world cells or scenarios. However, the state in this paper is based on the vector of features $\boldsymbol{\phi}(s)$ based on (\ref{eq:uav_feature_value}). Because the states are defined as continuous values from the feature vectors, this makes the Q-learning updating rule more difficult. The solution to this problem is to find the Q-value from the defined feature vectors using a linear function approximation (LFA) tool such as linear regression. This means each function of $Q(s, a)$ can be defined based on the values of features:
\begin{align}
\label{eq:Q_learning_LFA}
 Q(s_t, a_t) &= \theta_0 + \theta_1 \ \phi_1(s, a) + \theta_2 \ \phi_2(s, a) + \dots 
 \\ \nonumber
 & \qquad \ + \theta_\kappa \ \phi_\kappa(s, a)
 = \boldsymbol{\theta}^T \phi(s, a),
\end{align}
where $\kappa$ is the number of features and in this paper, there are five features for each visited state. $\theta_i$ are the weight vectors to find the value of $Q(s, a)$. To find the $\boldsymbol{\theta}$ vectors, we need to use a batch of sample data to update the values. To solve this linear regression problem, the stochastic gradient descent (SGD) approach is used to minimize the loss function for the difference between the predicted Q-value and the actual value. The true target value for the Q is obtained based on the Q-value updating rule and this value is called $Q^{+}(s, a)$:
\begin{align}
\label{eq:Q_learning_Q_target}
& Q^{+}(s, a) \approx R(s, a, s') + \gamma \max_{a'}Q(s', a')~. 
\end{align}
This Q target is used in the loss function for the SGD:
\begin{align}
\label{eq:SGD_loss}
L(\theta^{(i)}) &= \frac{1}{2}(Q^{+}(s, a) -  Q(s, a))^2 
\\ \nonumber
&= \frac{1}{2}(Q^{+}(s, a) -  \phi(s, a)^T \theta^{(i)})^2~.
\end{align}
Now considering the mentioned loss function, the values of the weights update based on the gradient:
\begin{align}
\label{eq:SGD_update_weight}
\theta^{(i+1)} &= \theta^{(i)} - \alpha_{S} \partial (L(\theta^{(i)}))/\partial \theta
\\ \nonumber
&= \theta^{(i)} - \alpha_{S} \partial (\frac{1}{2}(Q^{+}(s, a) -  \phi(s, a)^T \theta^{(i)})^2) / \partial \theta
\\ \nonumber 
&= \theta^{(i)} - \alpha_{S} [Q^{+}(s, a) -  \phi(s, a)^T \theta^{(i)}] \ . \ \phi(s, a) ~.
\end{align}

The optimal values of the weights are obtained based on different episodes, visits, and exploration of the agent. The agent's exploration and exploitation process is based on the $\epsilon$-greedy process with $\epsilon$ decay over the episodes. It means that the agent behaves randomly in the early episodes, and then acts more greedy based on exploitation of Q-values. At the end of this process, the learned policy ($\pi_i$) will be used in the apprenticeship learning to get the latest feature expectation values ($\mu_i$). Algorithm~\ref{algo:Q_Learning_LFA} summarizes the procedure for policy learning using Q-learning via linear function approximation for a specific weight of the reward function. 
\begin{algorithm}[hbtp]
\SetAlgoLined
 \textbf{Initialization:}\\
 Load the simulation environment\;
 $\epsilon = 1$\;
 \While{episode $<$ \textnormal{NUM\_EPS}}{
  $distance$ = 0\;
  Reset UAV()\;
  $Done$ = FALSE\;  
  \While{distance $<$ \textnormal{DIST\_LIM} \& not DONE}
  {
    \eIf{random $<$ $\epsilon$}
    {
        $action$ = random Action\;
    }
    {
        $action$ = GetGreedyAction(SGD Model)\;
    }
    Update $Location$ \;
    Calculate SNR, Interference, Throughput\;
    Calculate Features ($\phi_1(s, a, s'), \dots, \phi_5(s, a, s')$)\;
    Calculate immediate reward $R(s) = \boldsymbol{w}^T . \boldsymbol{\phi}(s')$\;
    $Q(s') = $ predict(SGD Model, $\phi(s')$)\;
    \eIf{New location is \textnormal{DESTINATION}}
    {
        $Done$ = TRUE\;
        $Q^+(s, a) = R(s)$\;
    }
    {
        $Q^+(s, a) = R(s) + \gamma \max\limits_{a'} Q(s', a')$\;
    }
    Update SGD Model(SGD Model, $\boldsymbol{\phi}(s)$, $Q^+(s, a)$)\;
    $distance += 1$\; 
  }
  \If{$\epsilon$ $>$ 0.1 \& episode $>$ \textnormal{NUM\_EPS/10}}{
  $\epsilon$ -= 1/\textnormal{NUM\_EPS} \;
  }
 $episode += 1$\;
 }
 \textbf{Return} SGD Model\;
 \caption{Q-learning algorithm with Linear Function Approximation}
 \label{algo:Q_Learning_LFA}
\end{algorithm}
Algorithm~\ref{algo:Q_Learning_LFA} finds and returns the learned policy for the specific reward function. $\textnormal{DIST\_LIM}$ indicate the UAV's hop counts limitation that based on UAV's battery life and flight time limitation is adopted. 
In this case, the learned policy is the trained SGD model. The size of the SGD model depends on the action vector. For instance, in this paper, the agent has 6 (Mobility) $\times$ 6 (Power Allocation) actions to choose from, hence 36 SGD models are the learned policy for this specific configuration.
In next section, we propose a deep Q-network instead of the stochastic gradient descent approach for linear function approximation.

\subsection{AL-IRL with Deep Q-Network}
\label{subsubsec:IRL_DQN}
\begin{figure}[bt]
	\centering
	\includegraphics[width=0.8\columnwidth]{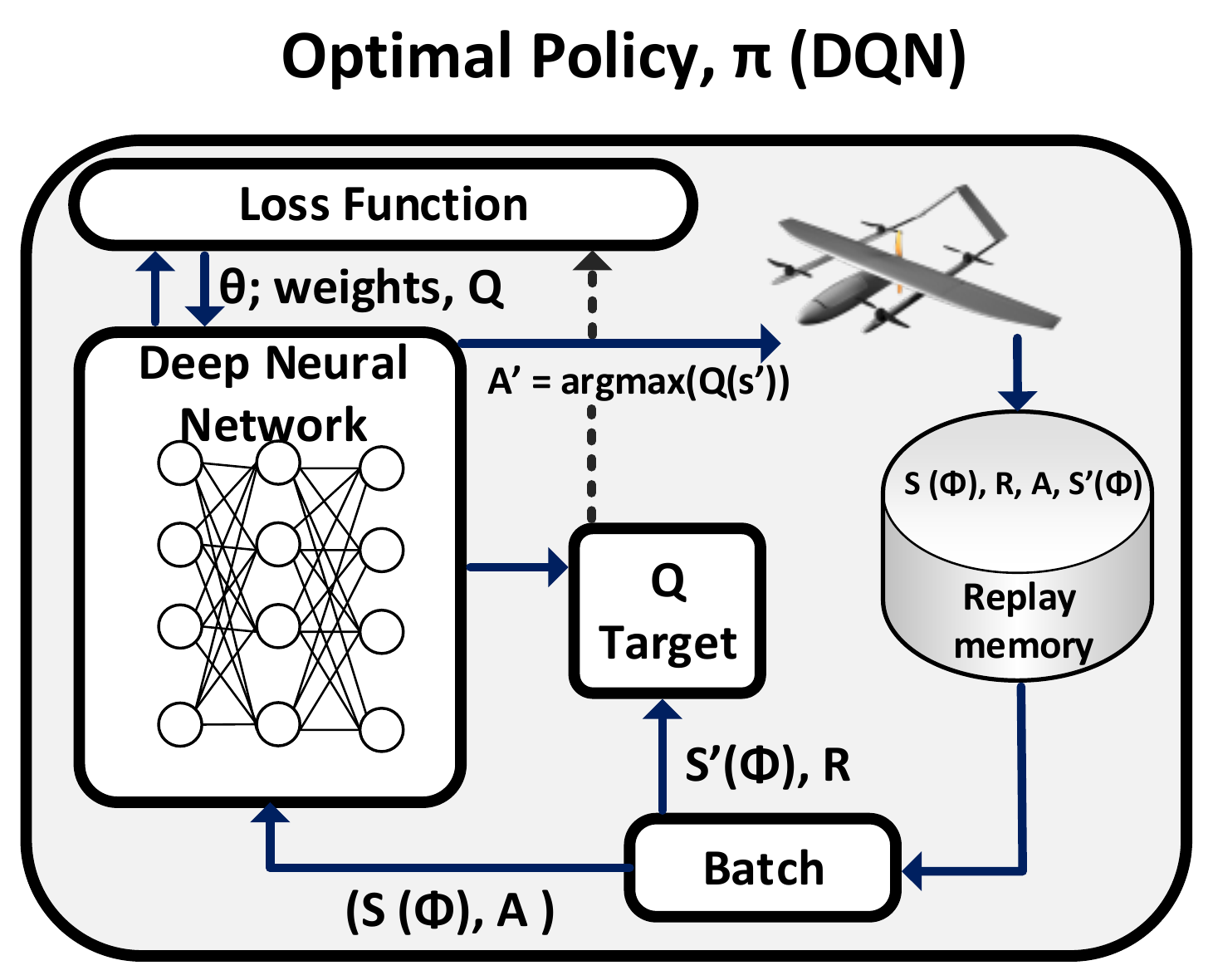}
	\caption{Deep Q-Network approach for the optimal policy in the apprenticeship learning.}
    \label{fig:AL_IRL_DQN}
\end{figure}

In the previous section, the solution to find the optimal policy was based on the Q-learning algorithm using the linear function approximation with stochastic gradient descent. In this section, the DQN approach is used to predict the Q-value based on the current features vector (current state) and the chosen action. Fig.~\ref{fig:AL_IRL_DQN} demonstrates the deep reinforcement learning structure for the apprenticeship learning for the obtained reward weights based on the inverse reinforcement learning algorithm. 

\begin{figure}[bt]
	\centering
	\includegraphics[width=1\columnwidth]{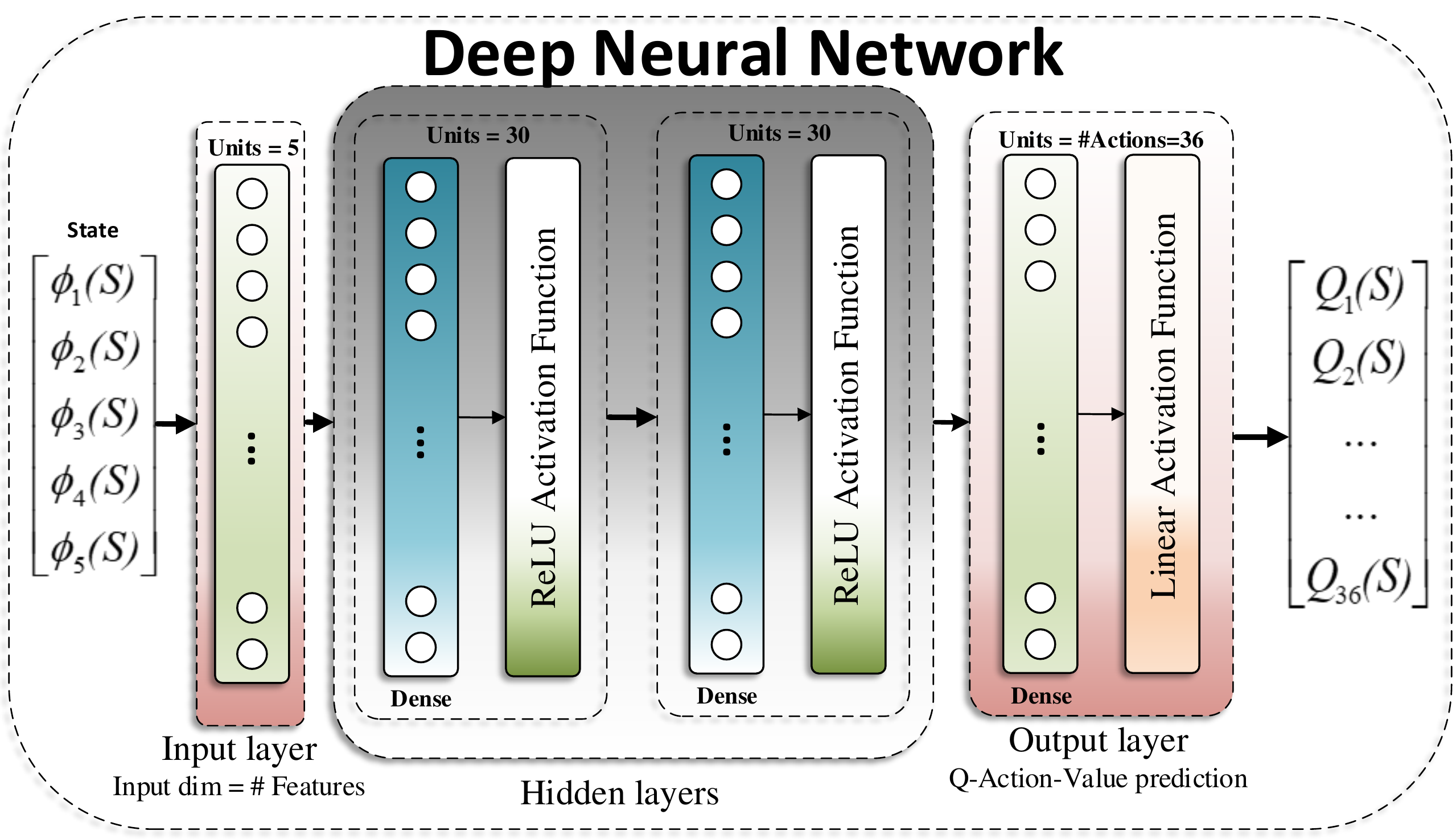}
	\caption{Deep neural network structure used in the DQN approach to predict the Q-Action value based on the features vector.}
    \label{fig:AL_IRL_DNN}
\end{figure}

The optimal policy learning block consists of a deep neural network with two hidden layers, an input layer with the size of the number of features for each state, and an output layer with the size of all actions. The numbers of neurons in the hidden layers are 30 and 30, respectively. Both hidden layers utilize the Rectified Linear Unit (RELU) \cite{li2017convergence} for the activation layer. The last layer has the linear activation function since the output of the deep neural networks is a continuous value for the Q-Action-Value. The structure of this DNN is demonstrated in Fig~\ref{fig:AL_IRL_DNN}.

The concept of the training for the exploration and exploitation is similar to Algorithm~\ref{algo:Q_Learning_LFA}. However, the training process is different and in this approach, training the model is based on the batch samples. The batch samples are picked randomly from the replay memory. Also, a buffer length is considered for the replay memory to keep the data fresh and pop the old data from the replay.
Algorithm~\ref{algo:dqn} presents the random batch sample training and fitting those data for the model. This algorithm is summarized since some structures appeared in Algorithm~\ref{algo:Q_Learning_LFA}, hence, a few lines are just removed. 

\begin{algorithm}[hbtp]
\SetAlgoLined
 \While{episode $<$ \textnormal{NUM\_EPS}}{
  ...\;
  \While{distance $<$ \textnormal{DIST\_LIM} \& not DONE}
  {
    ...\;
    Collect features, actions, reward for the Replay memory\;
    \If{episode $>$ \textnormal{NUM\_EPS/10}}{
        batch = random.sample(replay, BATCH\_SIZE) \;
        \For{all data in batch}
        {
            x\_train = batch[current feature]\;
            \eIf{New location is \textnormal{DESTINATION}}
            {
                $Q^+(s, a) = R(s)$\;
            }
            {
                $Q^+(s, a) = R(s) + \gamma \max\limits_{a'}Q(s', a')$\;
            }
            y\_train = batch[$Q^+(s, a)$]\;
        }
        model\_DQN.fit(x\_train, y\_train)\;
    }
    $\textnormal{distance} += 1$\; 
  }
  \If{$\epsilon$ $>$ 0.1 \& episode $>$ \textnormal{NUM\_EPS/10}}{
  $\epsilon$ -= 1/\textnormal{NUM\_EPS}\;
  }
 $episode += 1$\;
 }
 \textbf{Return} DQN Model\;
 \caption{Deep Q-Network to predict the Q-Action-Value on batch samples}
 \label{algo:dqn}
\end{algorithm}

To train the model and find the optimal weights for the neurons, the Mean Squared Error (MSE)~\cite{allen1971mean} is used as the loss function for the optimization. And to perform the stochastic optimization on the MSE, an optimizer called ``Adam'' is used to solve the problem for each batch of collected data \cite{kingma2014adam}. The MSE loss function for a batch with $n$ samples is shown below
\begin{align}
\label{eq:MSE_loss}
& MSE = \frac{\sum\limits_{i=1}^{n}(Q^+(s, a) - Q(s, a))^2}{n},
\end{align}
where $Q(s, a)$ is the predicted value based on the current features vector and $Q^+(s, a)$ is the actual values (target value) for the Q-value updated based on (\ref{eq:Q_learning_Q_target}). $n$ is the number of samples in each batch samples for the training. 

\subsection{Imitation Learning: Behavioral Cloning}
\label{subsec:BC}
This section proposes the model used in this paper for the BC. Usually, supervised learning tools can obtain the model for imitation learning to mimic the expert's behavior~\cite{torabi2018behavioral}. However, the BC approach only has the solution for those states observed by the expert. If the agent or UAV examines a state that the expert has never seen before, the outcome may not be optimal. All trajectories from the expert are gathered for imitation learning and BC. The same feature vectors of states with the respective actions are used to train a deep neural network for task classification based on the observed state by the UAV. This problem is categorized as a task classification problem since the UAV wants to choose the right action at a right time based on the expert's experience. 

To implement the supervised learning for the BC, we first collect the expert's data and trajectories. These trajectories provide a simulated dataset for the training and test evaluation. BC does not need constant access to the expert; however, DAGGER requires the expert's presence in states that have never been visited before. Implementing the DAGGER approach is more costly and not always feasible; thus, the BC is applied here for imitation learning. We should note that the collected dataset is based on states and actions, and there is no meaning of the reward function, and the agent mimics the expert blindly.  

The expert data, ($\mathcal{D}$), is collected in terms of $\mathcal{D} = \{(\mathbf{x}_t, \mathbf{s}_t, \mathbf{a}_t) \}$, where $\mathbf{x}_t$ is the initial condition for the expert. In this paper, we assume that $\mathbf{x}_t$ is the location where the agent or UAV starts its initial location $(0,\ 0)$. $\mathbf{s}_t$ is the UAV's state that consists of the features vector defined in (\ref{eq:uav_feature_value}) in Section~\ref{subsec:IRL}. Action ($\mathbf{a}_t $) is the mixed action based on the mobility and the transmission power allocation, $\mathbf{a} \in \mathcal{A}$. The action space is $\mathcal{A} = \{a(i) |  0\leq i \leq 35 \}$ which consists of 6 movement actions and 6 transmission powers. Since the agent decides at the same time for both the mobility and transmission power, there are 36 actions. If the actions were independent, the number of actions would be 12 instead of 36. The collected expert trajectories can be shown as,
\begin{align}\label{eq:expert_trajectories}
\tau = \{\mathbf{\phi(s)}^{(0)}, \mathbf{a}^{(0)}, \mathbf{\phi(s)}^{(1)}, \mathbf{a}^{(1)}, \dots \} \ .
\end{align}

After collecting the expert trajectories, the problem is defined as multi-class classification, and a supervised learning technique is utilized to handle the problem. The supervised learning approach generates a model predicting the appropriate class (action) based on the visited state. A decision tree algorithm is chosen as a supervised learning technique. Next, the UAV will use this model to decide on the next cell and the next transmission power based on the expert's collected information. Here, the learned model is called $\pi_{\rm BC}$
\begin{align}\label{eq:BC_model}
\mathbf{a} = \pi_{\rm BC}(\mathbf{x}_t, \mathbf{s}_t), 
\end{align}
which $\textnormal{BC}$ stands for the BC. To train the model, the decision tree from the SciKit-Learn Python package is used to solve the problem \cite{breiman1984classification, scikit-learn}. Gini impurity is used to determine the quality of split for various features. After training, the decision tree model has a depth of five with seven leaves.



\section{Numerical Results and Experiments}
\label{sec:Simulation}

In this section, first, the designed simulator is explained, then, the convergence of the apprenticeship learning for the IRL and the policy learner using both the Q-learning and deep Q-network methods is analyzed. Afterward, 
the performance comparison among various decision-making approaches at the UAV including inverse reinforcement learning (Q-learning using linear function approximation), inverse Deep Q-Network, BC, shortest path, and a random policy is utilized to compare the proposed method with similar approaches. Through all the experimental results a probabilistic channel according to \ref{eq:los_probability} for UAV is considered. However, in some cases, to show the performance comparison results in different UAV's channel conditions, we mentioned two different UAV's channel scenarios, probabilistic and LoS channels. All designs and simulations are tested on a Ubuntu system with a Ryzen 9 3900X CPU and Nvidia RTX 2080 Ti GPU. The training phase for the deep neural network approach was done based on Tensorflow~2.3.0, Keras~2.4.0 API, CUDA~10.1, and cuDNN~7.6 with Python~3.6.

\subsection{Designed Simulation Environment}
\label{subsec:sim_env}

In this study, we develop an Open Source simulation environment based on the proposed problem in Sec.~\ref{sec:System_Model_IRL}. The simulation environment is designed in Python 3.6 to consider the UAV in a pre-defined area with 25 cellular base stations and multiple terrestrial UEs.
The UEs' distribution is in such a way that some cells and areas have a higher density and others are less crowded.

The proposed solution should be able to find the optimal path regarding the distance, throughput, number of adjacent UEs, and interference on UEs in a dense urban environment with parameters $c_1=12.076$ and $c_2=0.114$. This simulation is event-based, and the event in this case is the UAV's action. 
The UAV has six actions for mobility and six discrete actions for the transmission power, which results in 36 actions together. The user can change the transmission power range to consider a broader or narrower range to increase or decrease the number of actions. More actions bring more complexity to convergence. 
\begin{figure}[bt]
	\centering
	\includegraphics[width=0.8\columnwidth]{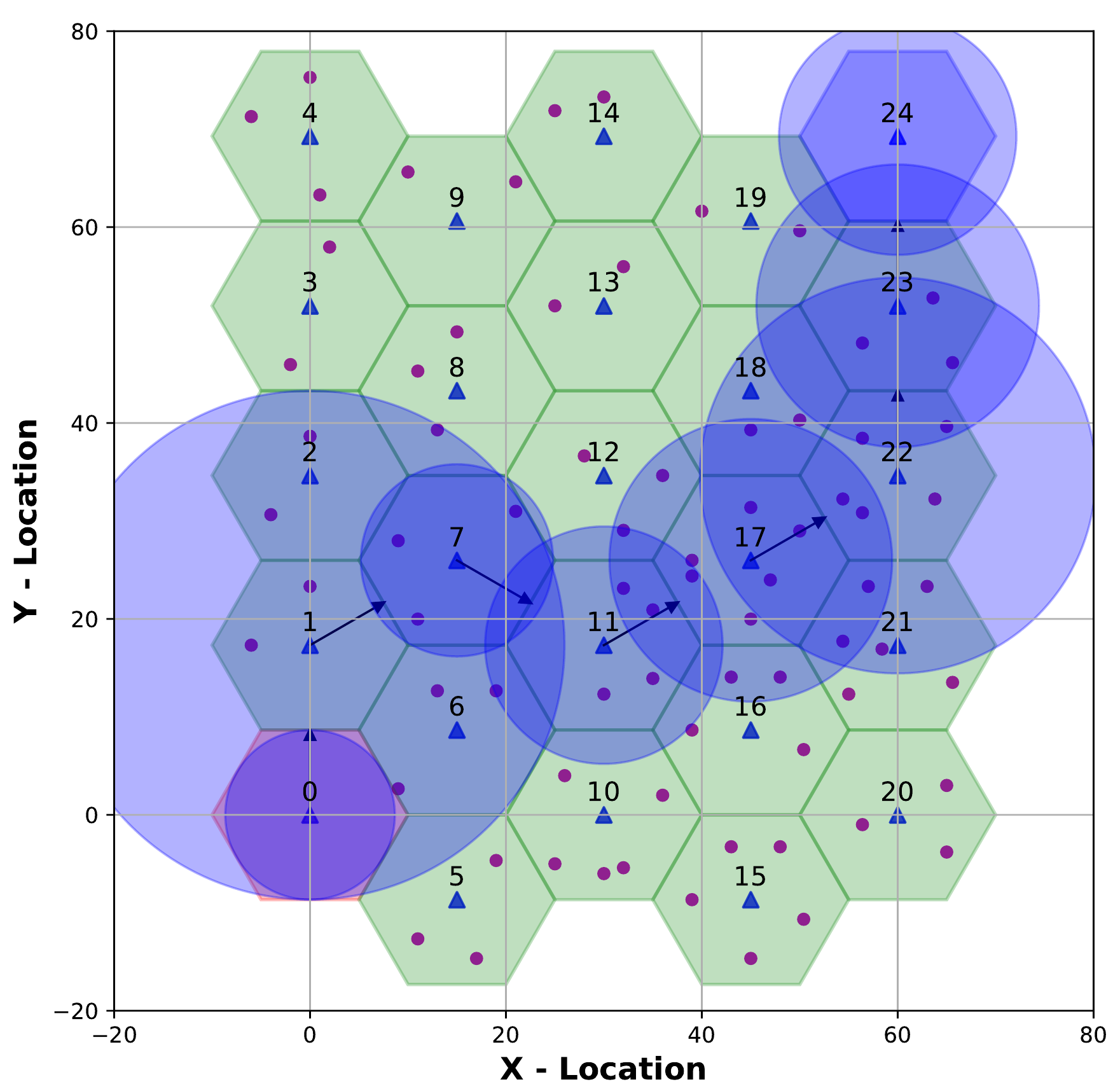}
	\caption{Demonstration of the designed graphical interface for the user.}
    \label{fig:simulator_IRL_UAV_cropped}
\end{figure}
This simulator is designed based on an object-oriented approach. Different classes are defined for the UAV, UEs, and BSs. All parameters for the learning algorithms and environment are configurable in the simulator. Table \ref{tab:parameter} summarizes the main simulation parameters.
\begin{table}[t!] 
	\footnotesize
	\centering
	\caption{\vspace*{-0cm} Simulation Parameters} \vspace{-0.0cm}
	\begin{tabular}{|>{\centering\arraybackslash}m{3.5cm}|>{\centering\arraybackslash}m{3cm}|}
		\hline 
		\vspace{0.2cm}\bf{Parameter} &\vspace{0.2cm}\bf{Value } \\
		\hline \vspace{0.2cm}
		Number of cellular BS & $25$ \\
		\hline \vspace{0.2cm}
		Number of UEs & $75$   \\
		\hline \vspace{0.2cm}
		UAV altitude ($H$) & $50$ m  \\
		\hline \vspace{0.2cm}
		UAV max transmit power  & $200$ mW\\
		\hline\vspace{0.2cm}
		UAV min transmit power  & $50$ mW\\
		\hline\vspace{0.2cm}
		Dense urban ($\eta_{LoS}$, $\eta_{NLoS}$)  & $(1.6,23)$ [dB] \\ 
		\hline\vspace{0.2cm}
		Dense urban ($c_1$, $c_2$)  & $(12.076,0.114)$ \\ 
		\hline\vspace{0.2cm}
		$N_0$ & $-90$ dBm \\ 
		\hline\vspace{0.2cm}
		Carrier frequency & $2$ GHz \\ 
		\hline\vspace{0.2cm}
		UE transmit power & $2$ mW \\
		\hline\vspace{0.2cm}
		UAV antenna gain & $100$ \\
		\hline\vspace{0.2cm}
		Number of features & $5$ \\
		\hline\vspace{0.2cm}
		Number of Epochs & 1e4 \\
		\hline\vspace{0.2cm}
		$\epsilon$ & $0.1$ \\
		\hline\vspace{0.2cm}
		Batch size & $24$ \\
		\hline\vspace{0.2cm}
		Replay buffer size & $1e4$  \\
		\hline\vspace{0.2cm}
		$\gamma$ & $0.99$ \\
		\hline\vspace{0.2cm}
		$\alpha$ & $0.001$ \\
		\hline
	\end{tabular}\label{tab:parameter} \vspace{-0cm}
\end{table}

In Fig. \ref{fig:simulator_IRL_UAV_cropped}, a circle notation is considered to show the UAV's coverage area as a function of its transmission power for the sake of demonstrating the interference level. The UAV's directions are demonstrated by black arrows to show the chosen path. The user can turn off the Graphical User Interface (GUI) to speed up the simulation. All models and trajectories can be saved on a drive for future evaluations. Fig.~\ref{fig:simulator_IRL_UAV_cropped} shows a sample snapshot of the simulator with the chosen path. In this sample, the UAV flies from the source cell (BS$_0$) toward the destination cell (BS$_{24}$) with varying transmission power. The source code of this simulator is available on the GitHub repository \cite{github:code_AL_IRL2021}. A sample video of the designed simulator implementing the different approaches in this study is available on YouTube~\cite{youtube2021_inverserl}. In the video, we showed the simulation based on the different approaches during the training phase and also after training for the evaluation. 

\subsection{Convergence of the IRL}
\label{subsec:sim_converge_IRL}

\begin{figure}[bt]
	\centering
	\includegraphics[width=0.8\columnwidth,keepaspectratio]{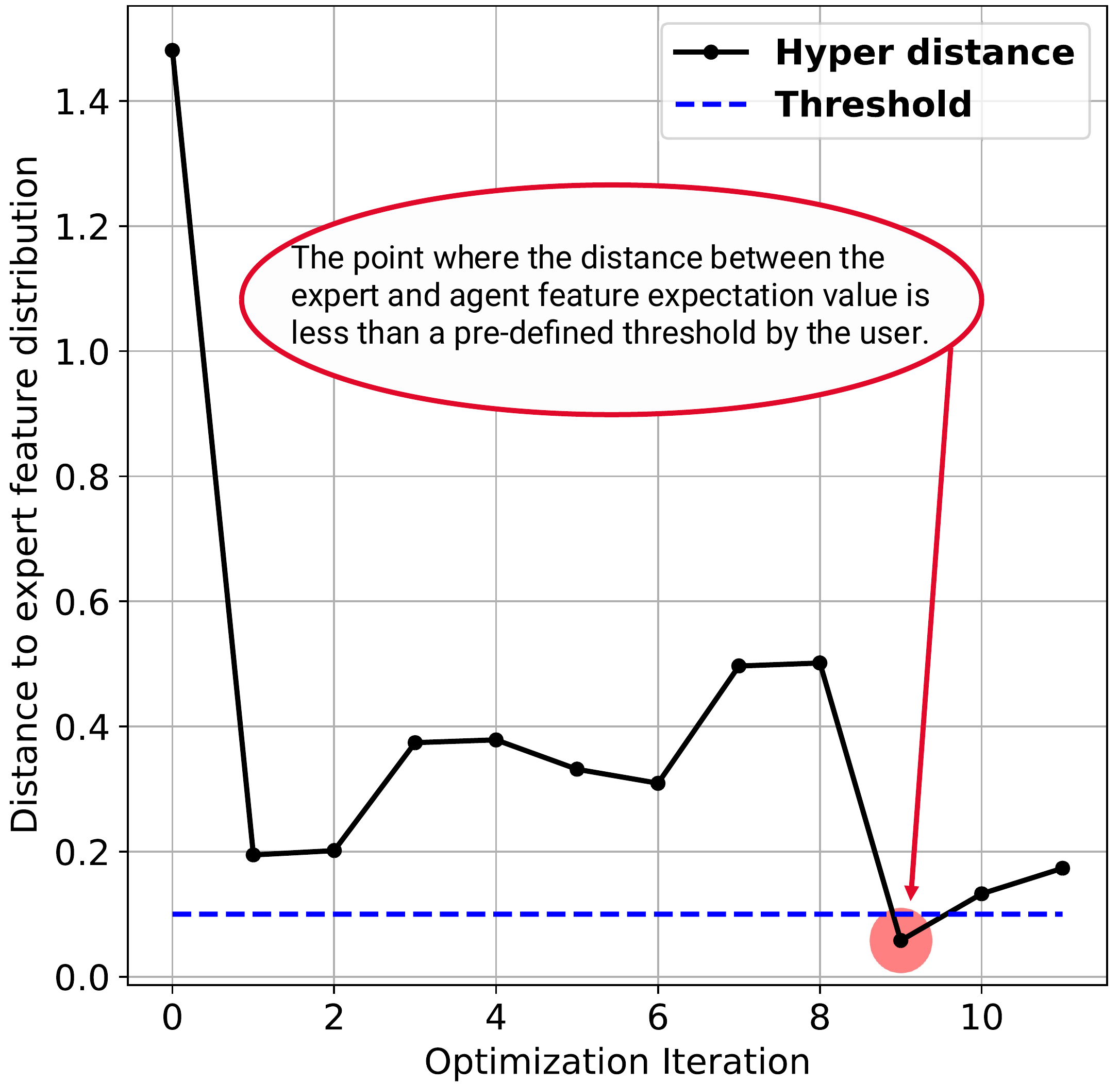}
	\caption{Hyper distance between the expert and agent feature expectation based on different weights, reward functions, and optimal policy.}
    \label{fig:Hyper_distance}
\end{figure}

To investigate the convergence for Inverse Reinforcement Learning, the hyper distance between the expert's feature expectation and learner feature expectation is considered as a metric to show the convergence behavior of the IRL algorithm. 
Fig.~\ref{fig:Hyper_distance} demonstrates this hyper distance for different iterations. The learning process stops at the iteration where the hyper-distance is getting lower than a predefined threshold ($\epsilon_{\textnormal{IRL}}$). The user can choose the desired threshold based on the requirements. If the user chooses $\epsilon_{\textnormal{IRL}} = 0$, this means the user wants to achieve the same behavior compared to the expert. Choosing $\epsilon_{\textnormal{IRL}} = 0$ may need an infinite number of iterations for the IRL algorithm to find the optimal weight. 

In Fig.~\ref{fig:Hyper_distance}, we assumed that the threshold ($\epsilon_{\textnormal{IRL}}$) for the distance between the expert and agent feature expectation is $0.1$ and any distance less than the threshold stops the algorithm. After ten iterations, this algorithm stops when the distance is $0.057$.

\subsection{Convergence of the Q-Learning and Deep Q-Network}
\label{subsec:sim_converge_QL_DQN}
\begin{figure*}
    \centering
    
    \begin{subfigure}{0.32\linewidth}
        \centering
        \includegraphics[width=\textwidth]{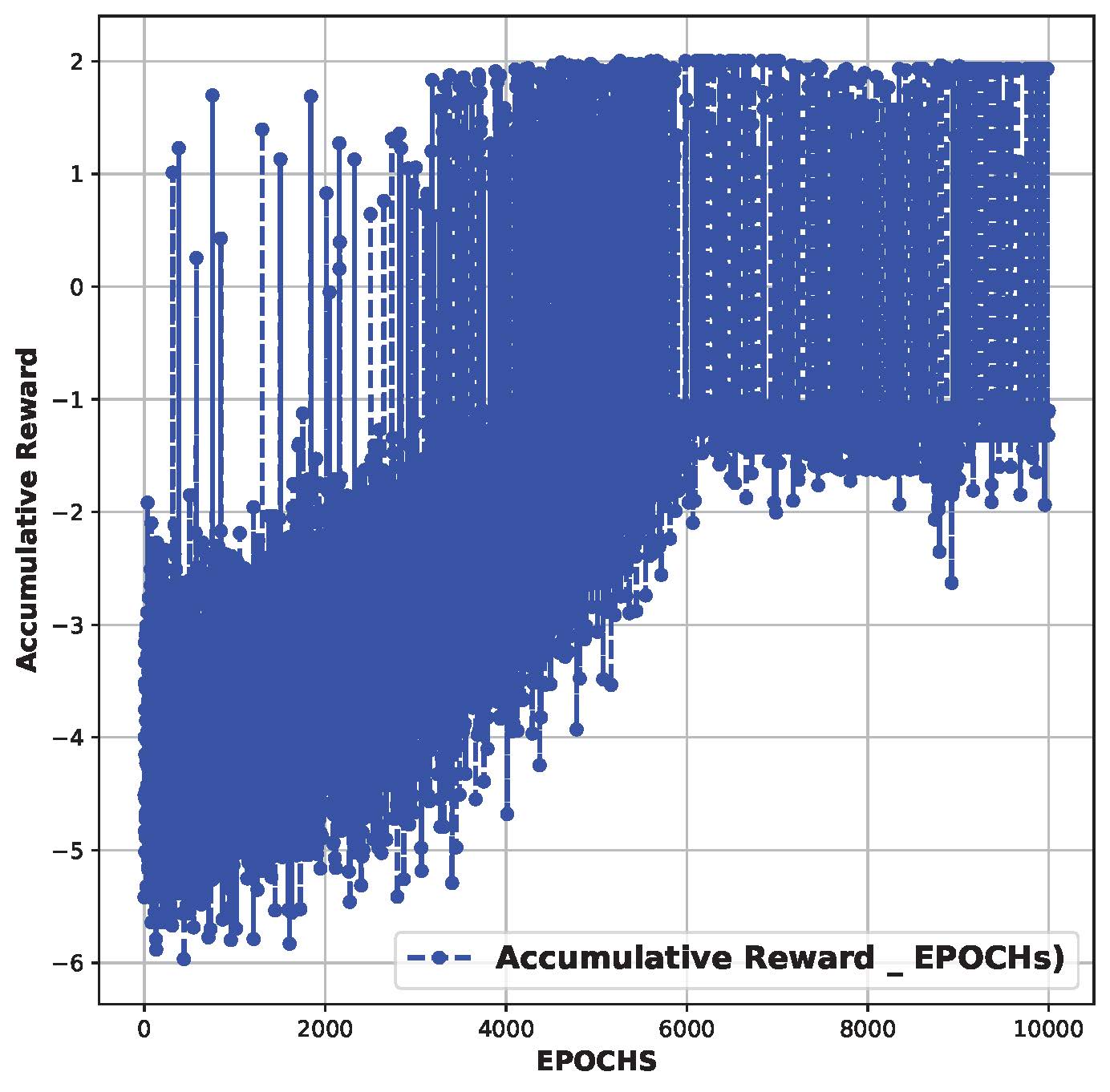}
        \caption{Accumulative reward in each episode at iteration 1}
        \label{subfig:qrl_convergence_it1}
    \end{subfigure}
    \hfill
    \begin{subfigure}{0.32\linewidth}
        \centering
        \includegraphics[width=\textwidth]{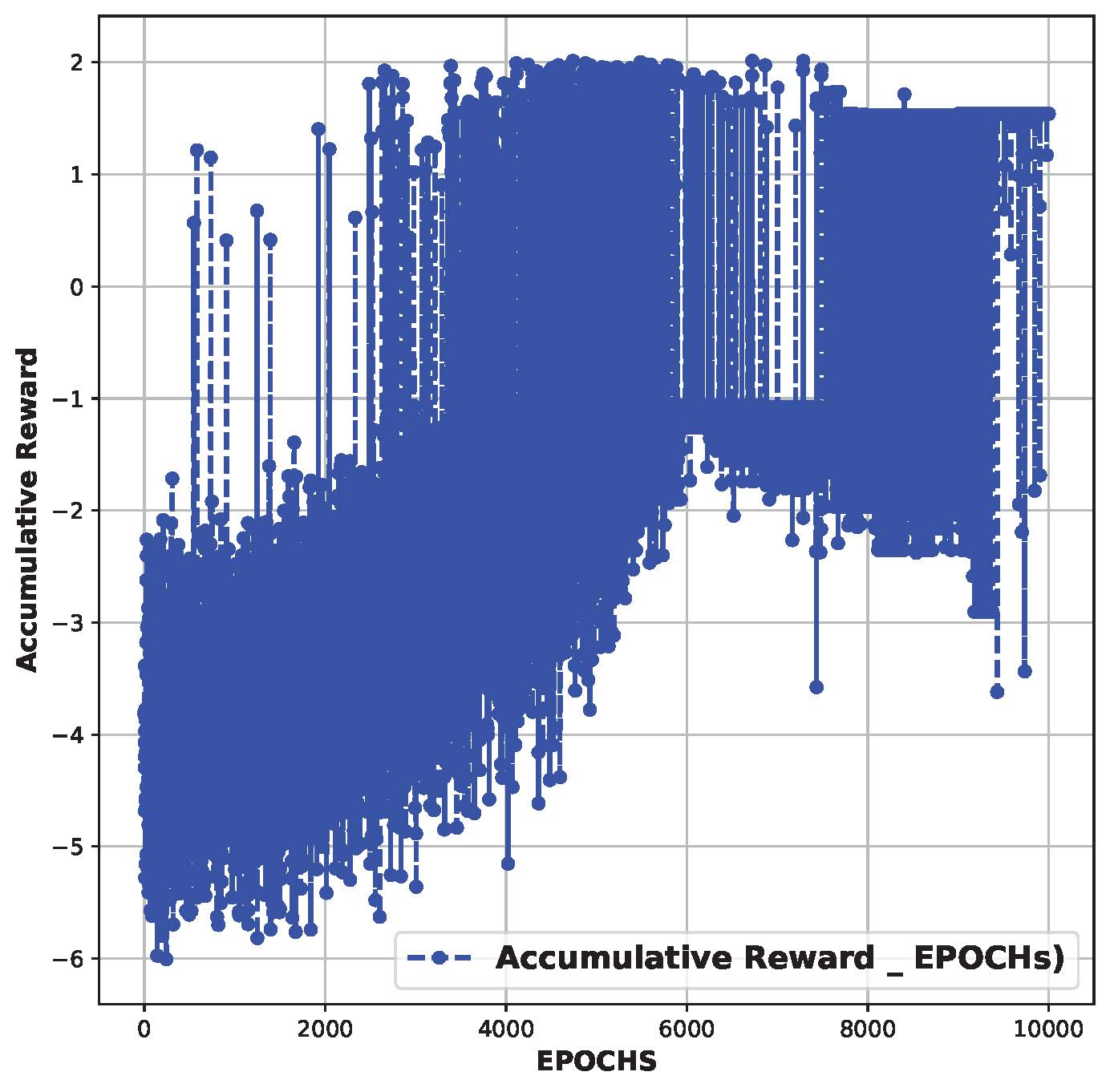} 
        \caption{Accumulative reward in each episode at iteration 2}
        \label{subfig:qrl_convergence_it3}
    \end{subfigure}
    \hfill
    \begin{subfigure}{0.32\linewidth}
        \centering
        \includegraphics[width=\textwidth]{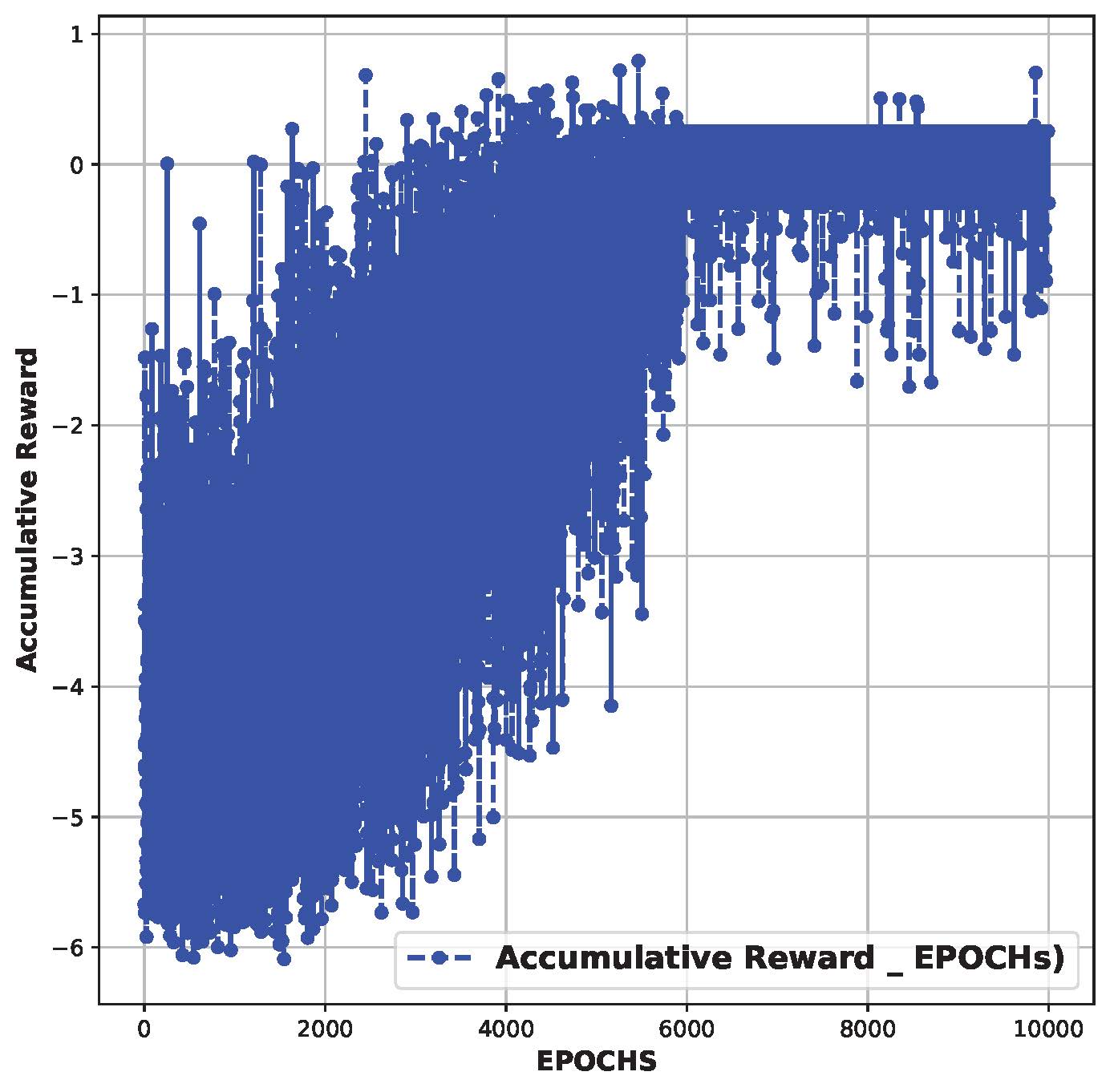}
        \caption{Accumulative reward in each episode at iteration 3}
        \label{subfig:qrl_convergence_it4}
    \end{subfigure}

    \caption{Convergence of the policy learning using Q-Learning in different iterations for the IRL approach.}
    \label{fig:Q-learning_reward_convergence}
\end{figure*}

\begin{figure*}
    \centering
    
    \begin{subfigure}{0.32\linewidth}
        \centering
        \includegraphics[width=\textwidth]{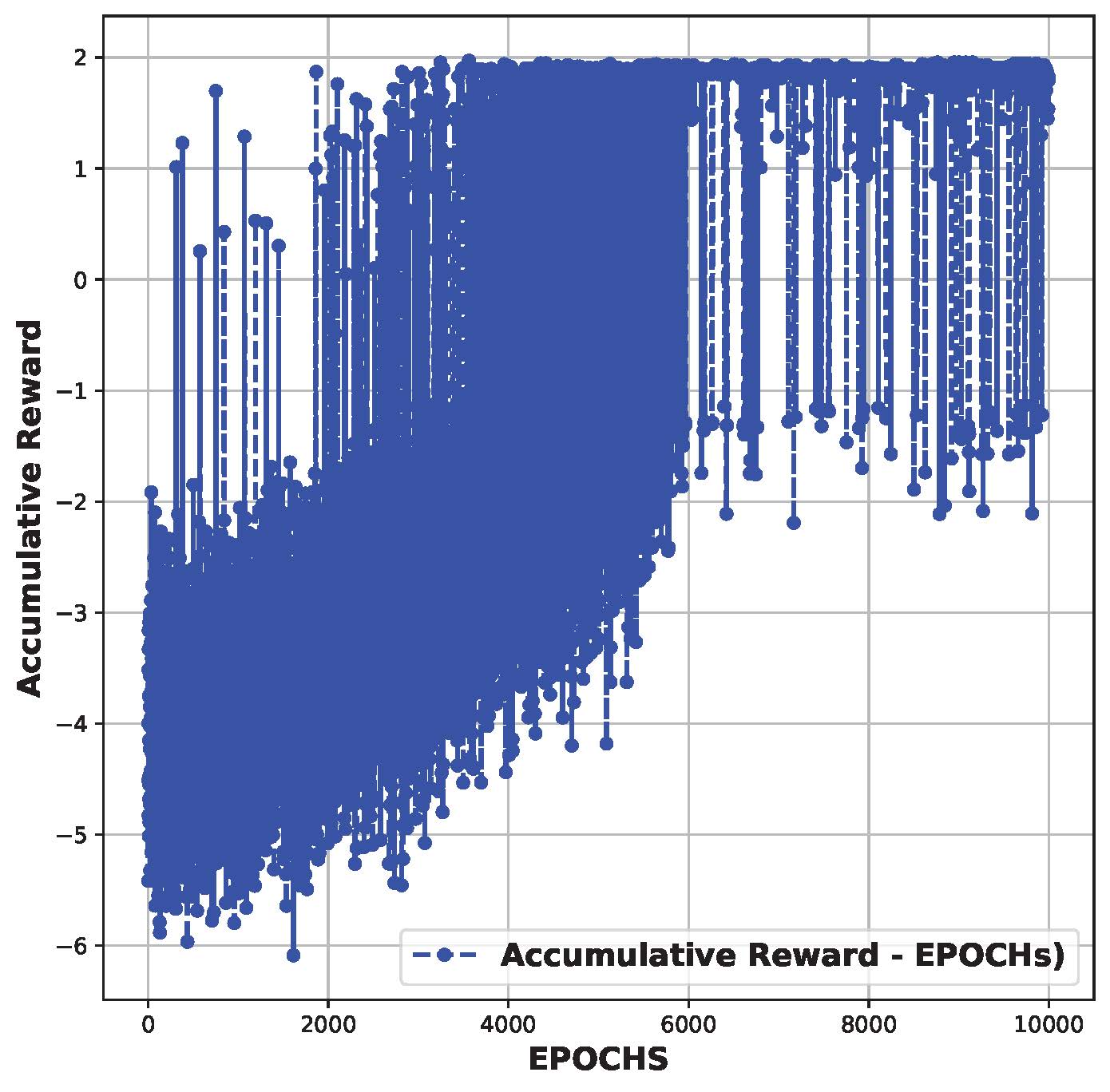}
        \caption{Accumulative reward in each episode at iteration 1}
        \label{subfig:dqn_convergence_it1}
    \end{subfigure}
    \hfill
    \begin{subfigure}{0.32\linewidth}
        \centering
        \includegraphics[width=\textwidth]{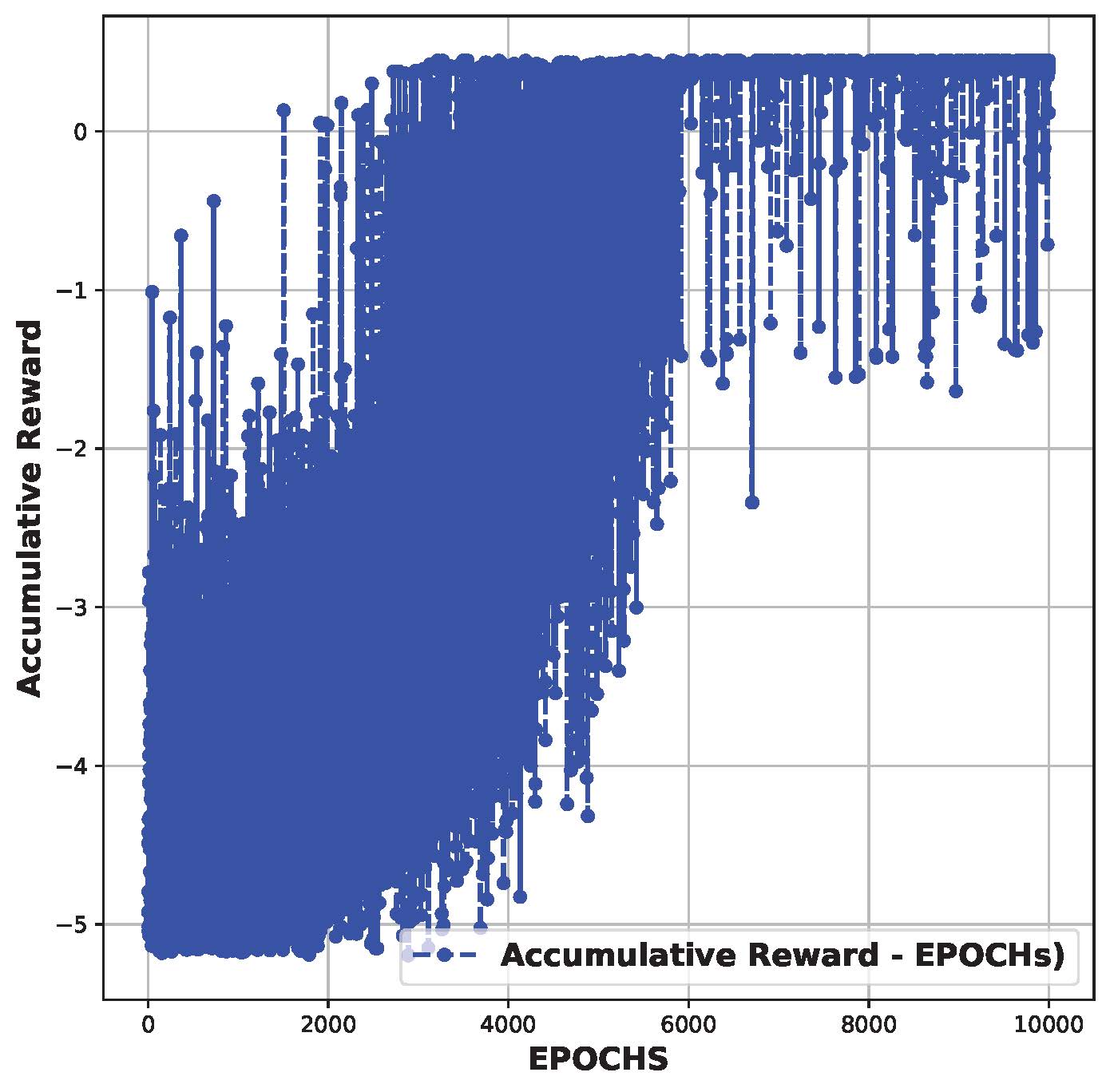}
        \caption{Accumulative reward in each episode at iteration 2}
        \label{subfig:dqn_convergence_it2}
    \end{subfigure}
    \hfill
    \begin{subfigure}{0.32\linewidth}
        \centering
        \includegraphics[width=\textwidth]{Figures/n_DQN_accumulative_reward_training_Feature_5_learner_1_EPOCHS_10000.jpg}
        \caption{Accumulative reward in each episode at iteration 3}
        \label{subfig:dqn_convergence_it3}
    \end{subfigure}

    \caption{Convergence of the policy learning using deep Q-Network in different iterations for the IRL approach.}
    \label{fig:DQN_reward_convergence}
\end{figure*}

Fig.~\ref{fig:Q-learning_reward_convergence} demonstrates a few iterations of the feature policy learning for each unique reward function for the Q-learning with linear function approximation using the SGD. Based on the exploration-greedy rate, the accumulative reward function is converged to the best optimum value. It is worth mentioning that the optimum value for the reward function depends on the obtained weight for the reward function at each iteration based on the QP, and because of that, each plot is converged to a different value. Also, Fig.~\ref{fig:DQN_reward_convergence} illustrates the same convergence concept for the various iterations of the policy learning for the Deep Q-Network. 

\subsection{Apprenticeship Learning via Inverse RL performance - Training Phase}
\label{subsec:sim_AL_IRL_performance_training}
\begin{figure*}[t]
\centering
\begin{subfigure}{0.32\textwidth}
\includegraphics[width=1\columnwidth]{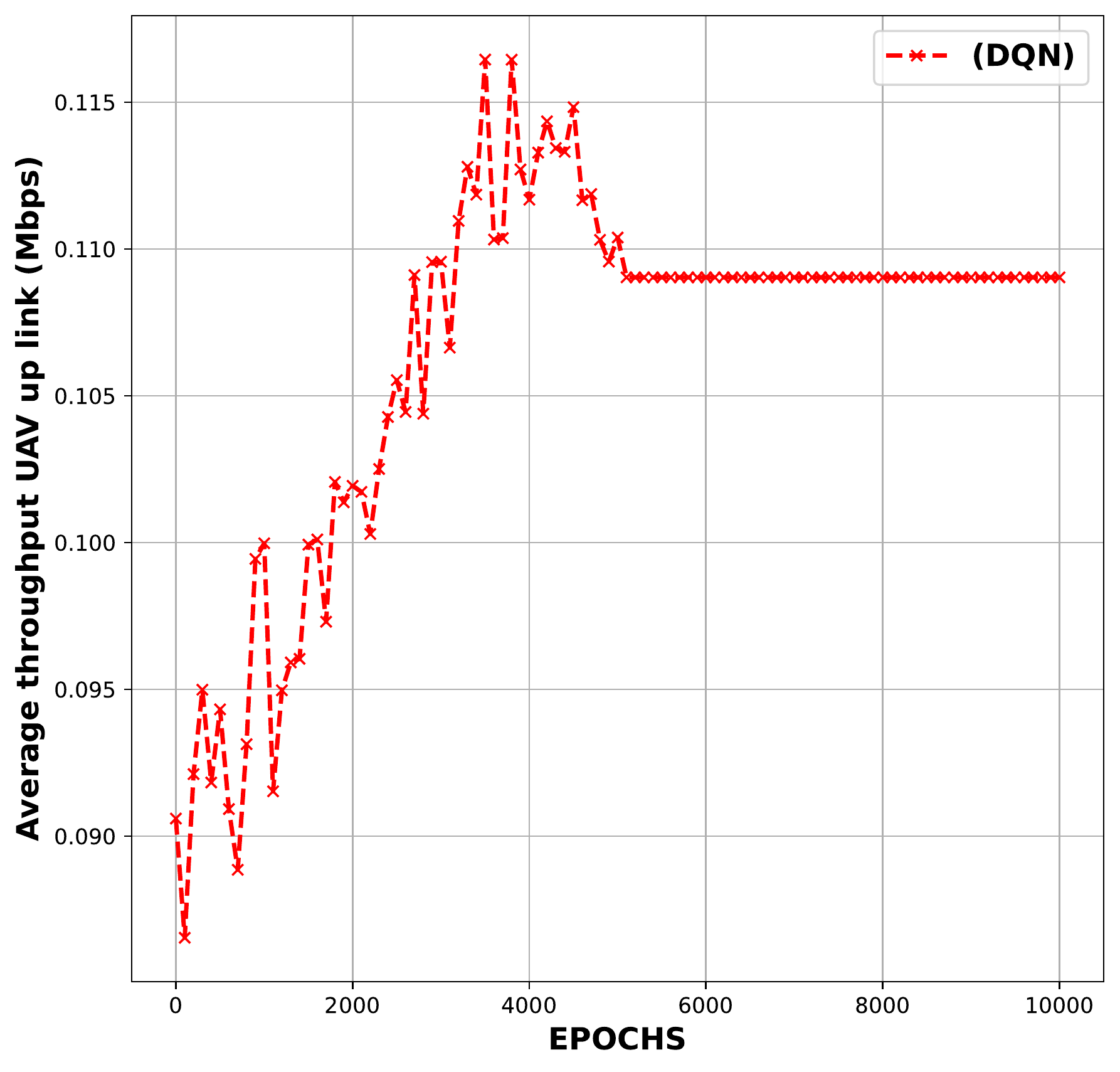}
	\caption{Transmission throughput rate for the UAV's Up-Link in probabilistic channel scenario}
    \label{fig:Throughput_learning_epochs_101}
\end{subfigure}
\hfill
\begin{subfigure}{0.32\textwidth}
\includegraphics[width=1\columnwidth]{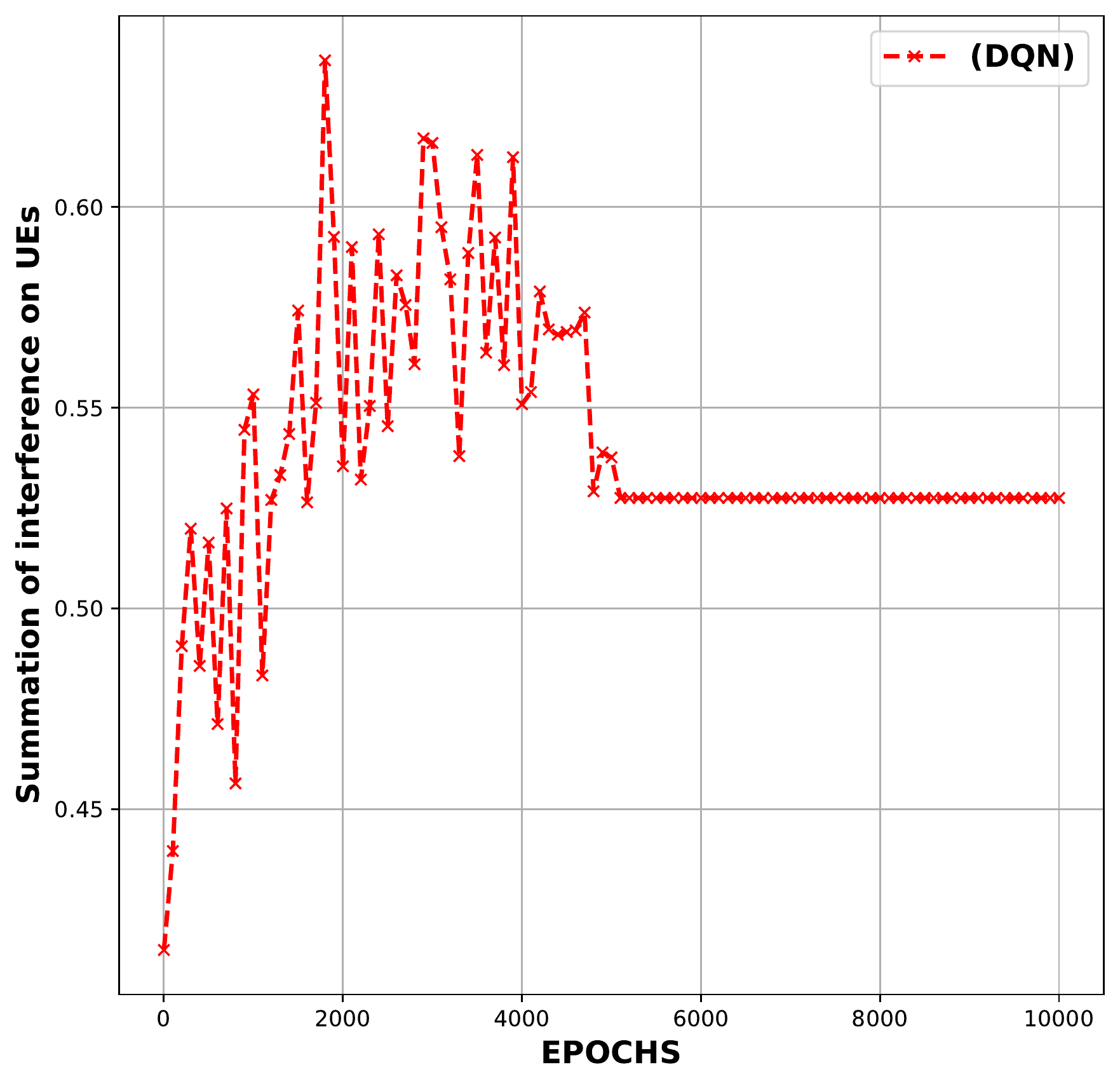}
	\caption{UAV's UL interference on neighbor UEs using the same RBs in probabilistic channel scenario}
    \label{fig:Interference_learning_epochs_101}
\end{subfigure}
\hfill
\begin{subfigure}{0.32\textwidth}
\includegraphics[width=1\columnwidth]{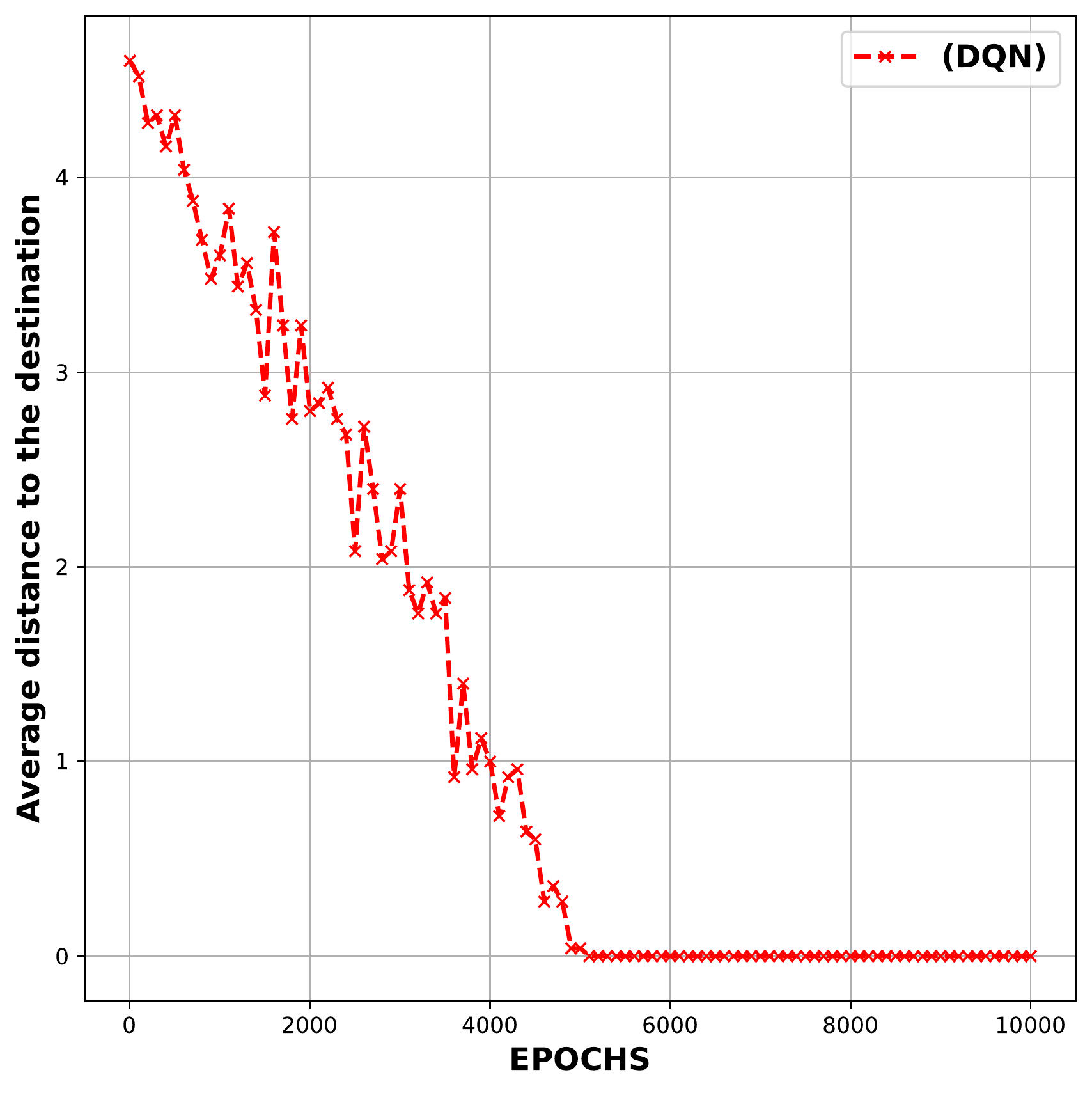}
	\caption{The distance between the last location where the UAV stops its task in probabilistic channel scenario}
    \label{fig:Distance_learning_epochs_101}
\end{subfigure}
\hfill
\begin{subfigure}{0.32\textwidth}
\includegraphics[width=1\columnwidth]{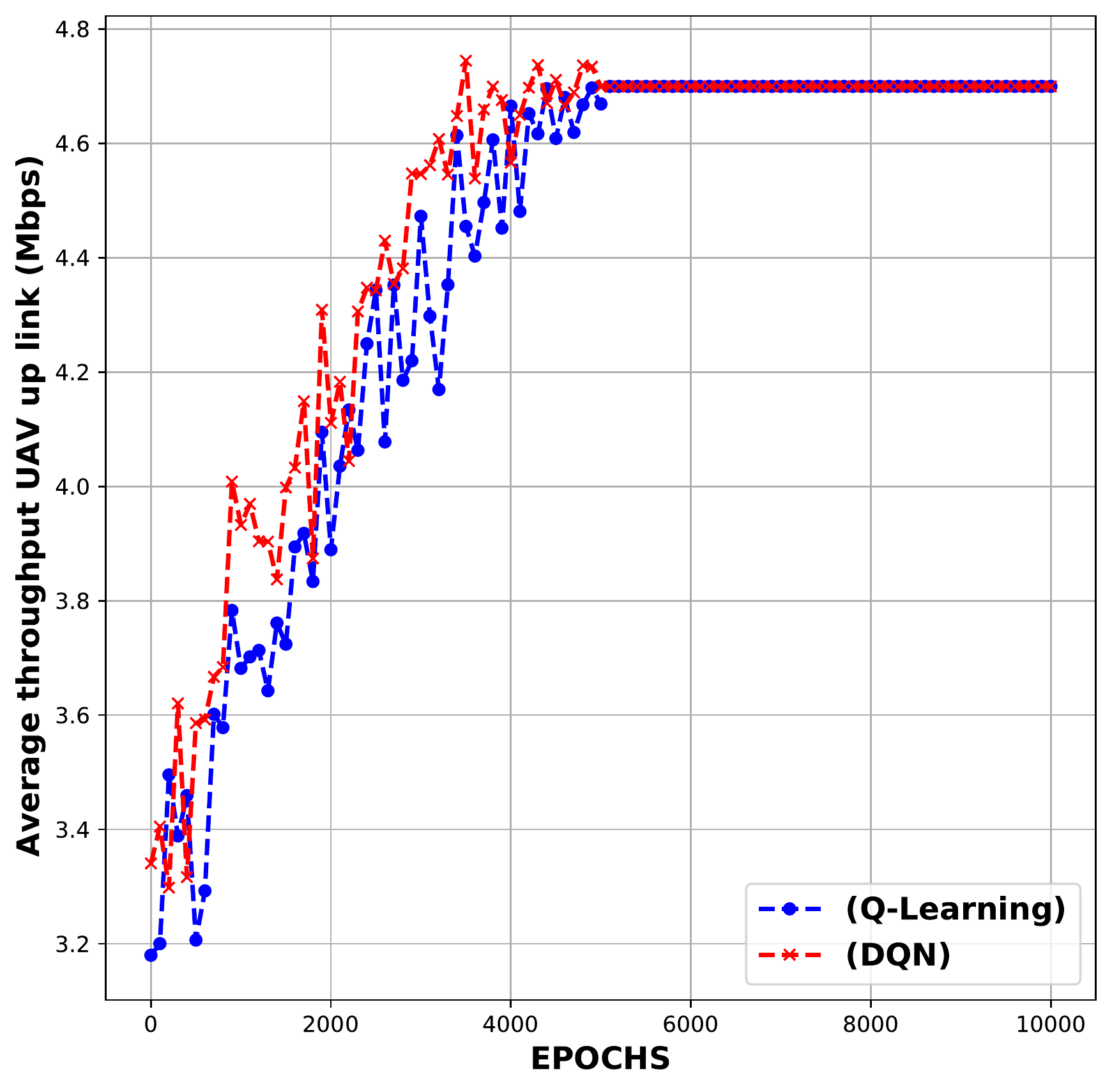}
	\caption{Transmission throughput rate for the UAV's Up-Link in LoS channel scenario}
    \label{fig:Throughput_learning_epochs_101_los}
\end{subfigure}
\hfill
\begin{subfigure}{0.32\textwidth}
\includegraphics[width=1\columnwidth]{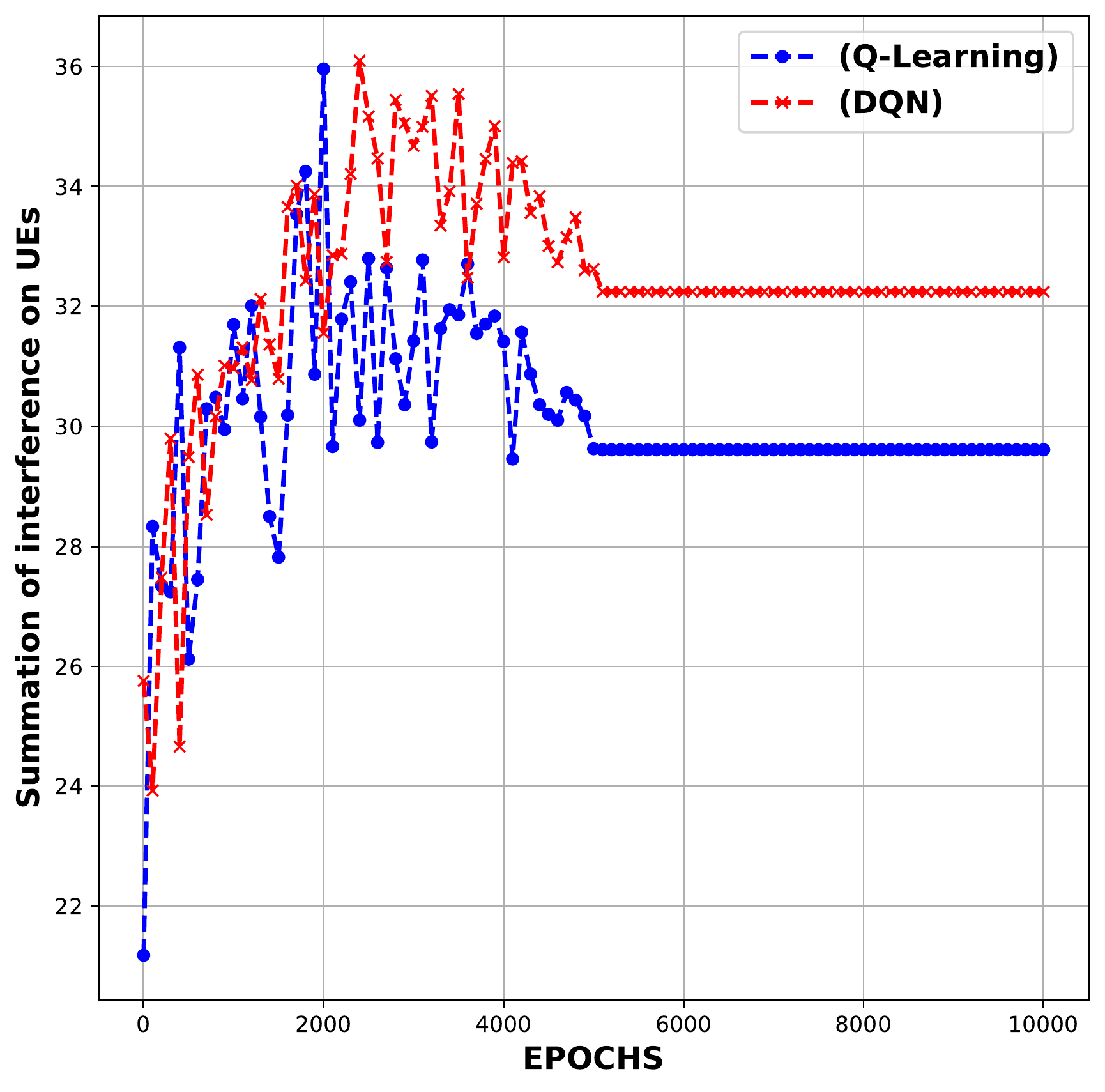}
	\caption{UAV's UL interference on neighbor UEs using the same RBs in LoS channel scenario}
    \label{fig:Interference_learning_epochs_101_los}
\end{subfigure}
\hfill
\begin{subfigure}{0.32\textwidth}
\includegraphics[width=1\columnwidth]{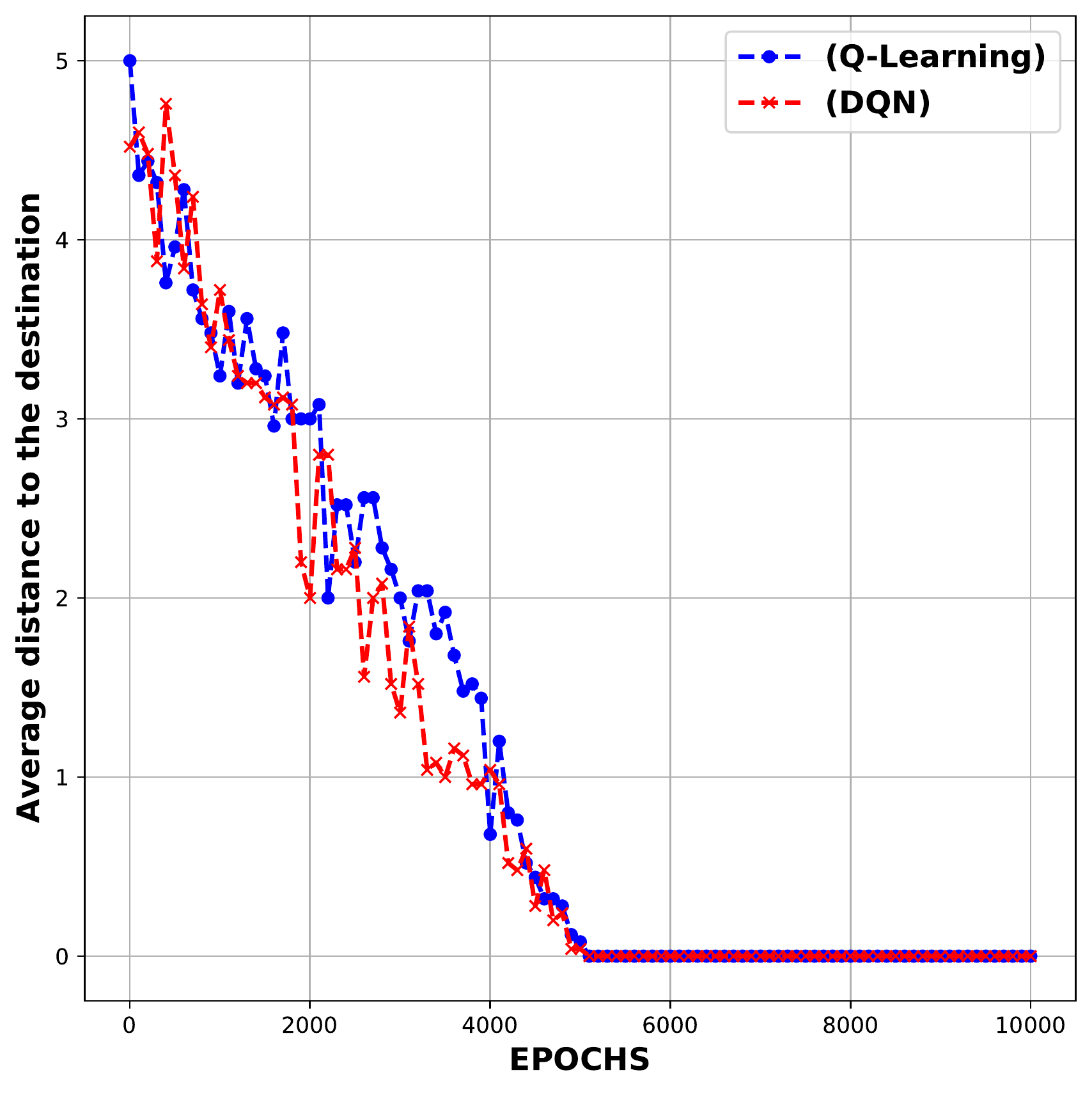}
	\caption{The distance between the last location where the UAV stops its task in LoS channel scenario}
    \label{fig:Distance_learning_epochs_101_los}
\end{subfigure}
\caption{Evaluation of UAV's UL throughput, interference value on neighbor UEs, and the final distance of the UAV during the training phase of the inverse RL for both Q-learning and DQN.}
\end{figure*}

All results in this section are averaged over 25 runs to have a smooth and clear demonstration for the sake of comparison. 
Figures~\ref{fig:Throughput_learning_epochs_101},~\ref{fig:Interference_learning_epochs_101},~\ref{fig:Distance_learning_epochs_101} represent results for UAV's probabilistic channels, and Figures ~\ref{fig:Throughput_learning_epochs_101_los},~\ref{fig:Interference_learning_epochs_101_los},~\ref{fig:Distance_learning_epochs_101_los} show the results for UAV's LoS channels.  
Fig.~\ref{fig:Throughput_learning_epochs_101} and Fig.~\ref{fig:Throughput_learning_epochs_101_los} show the throughput result of the inverse RL using the Q-learning and deep Q-network during the training phase of the optimal reward function after the final optimization and termination of the IRL algorithm. 
In Fig.~\ref{fig:Throughput_learning_epochs_101_los} both approaches converge at the same number of epochs and at the same values after the $\epsilon$-greedy exploration. It is worth mentioning that achieving the maximum throughput is not always the optimum policy since it increases interference as well. 


Also, Fig.~\ref{fig:Interference_learning_epochs_101} and Fig.~\ref{fig:Interference_learning_epochs_101_los} show the summation of interference levels on the neighbor UEs when the UAV uses the same resource block for its transmission and the UAV's uplink can interfere with the other UEs' downlink as well. Keeping the interference as low as possible is desired; however, it decreases the throughput as well, and it is not in line with the problem objectives defined in (\ref{eq:obj_main}) and (\ref{eq:obj_1}) in Section~\ref{sec:System_Model_IRL}. As Fig.~\ref{fig:Interference_learning_epochs_101_los} represents Q-learning and DQN algorithms converge to different values, and the reason is that because these two algorithms utilized different paths for the problem, thus the agent (UAV) senses a different number of UEs, and the effect of its UL transmission is different in these models. 


Fig.~\ref{fig:Distance_learning_epochs_101} and Fig.~\ref{fig:Distance_learning_epochs_101_los} demonstrate the distance between the location where the UAV finishes its task and the destination cell where it was supposed to finish and stop based on resource limitations such as the battery capacity and flight time. In Fig.~\ref{fig:Distance_learning_epochs_101} DQN converges to the distance of zero which means this algorithm finds the destination cell (in this scenario (BS$_{24}$)).
However, the limited energy of UAVs for flight, coupled with the highly dynamic wireless environment and the low convergence power of the Q-learning algorithm, results in an inability to reach the destination in probabilistic channel conditions. 
However, in Fig.~\ref{fig:Distance_learning_epochs_101_los} both Q-learning
and DQN converges to the distance of zero which means
both algorithms find the destination cell. 


\subsection{Apprenticeship Learning via Inverse RL Performance}
\label{subsec:sim_AL_IRL_performance}
Here, the goal is to compare the performance of different path planning mechanisms including apprenticeship learning via IRL, inverse deep RL, BC, shortest path with minimum yaw, pitch, and roll, and random action. The previous section shows the result for the inverse RL using Q-learning and DQN during the training phase. Here, based on the trained model and obtained weights for the reward function, a sample scenario is considered to compare the performance of these different approaches. The implemented BC model defined in Section~\ref{subsec:BC} predicts the desired action based on the visited features vector state with 89.08\% accuracy using the decision tree classification approach. 

\begin{figure}
\centering
\begin{subfigure}{0.7\linewidth}
\includegraphics[width=1\columnwidth]{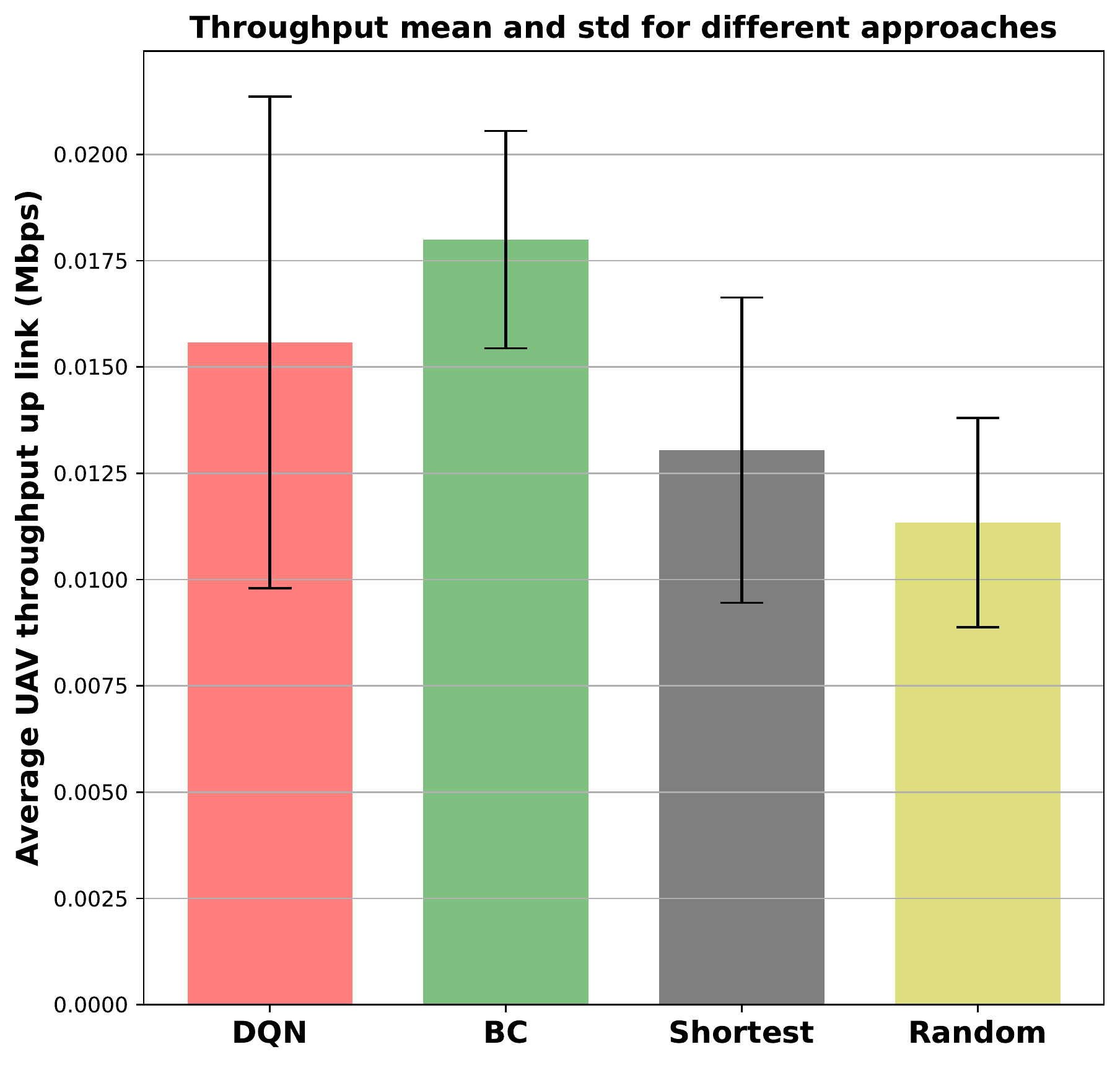}
	\caption{Throughput mean and standard deviation comparison for the UAV Up-link}
    \label{fig:Throughput_sample_steps_7}
\end{subfigure}
\vfill
\begin{subfigure}{0.7\linewidth}
\includegraphics[width=1\columnwidth]{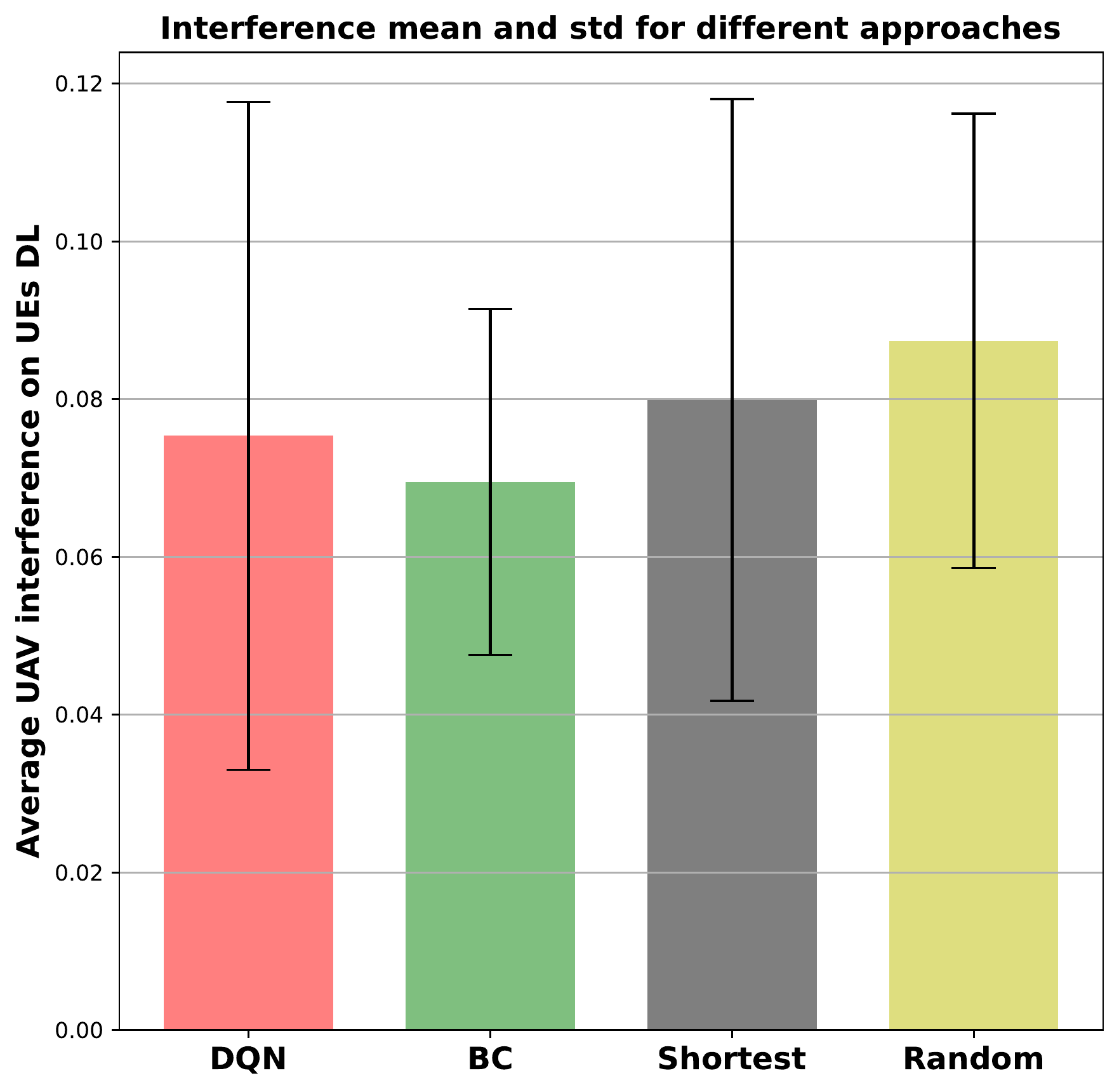}
	\caption{Average and standard deviation interference affecting the ground neighbor UEs}
    \label{fig:Interference_sample_steps_7}
\end{subfigure}
\caption{Evaluation of the UAV's UL throughput and the effect of interference with neighbor UEs in a single scenario.}
\end{figure}

Fig.~\ref{fig:Throughput_sample_steps_7} compares the throughput of the DQN, BC, shortest path, and random policy. DQN has better behavior for the throughput than random and shortest path algorithms, and the reason is the feature expectation for the throughput metric has a positive role on the reward function in the features state vector ($\phi(s, t)$). In the shortest path, the UAV chooses the shortest path between the source cell (BS$_0$) and the destination with minimum yaw, pitch, and rolling to consume less energy. The transmission power is chosen at random in the shortest path. The movement action and transmission power are both random in the random policy. Random policy has the worst performance in terms of throughput in the same scenario. The main factor to affect the throughput metrics is the agent's (UAV) transmission power for its UL transmission. \vspace{-0.1cm}


Another comparison is performed in Fig.~\ref{fig:Interference_sample_steps_7} regarding the interference management to compare all approaches. This interference is the summation of the interference applied to all neighbor UEs if they all use the same resource block. The level of interference in this figure depends on the UAV's transmission power for its UL, the density of neighbor UEs in the neighbor cells, and the distance between the UAV and the affected UEs. BC has the lowest interference level since it completely mimics the expert's behavior without knowing any understanding of it. Next, DQN approache has the next level for applied interference. The shortest path chooses the path that has a higher density of UEs density compared to other approaches, and because of that, the interference is higher compared to other two techniques and almost similar to random policy. If the user wants to train the DQN approache closer to the expert's behavior, then it is possible to choose a lower value for the epsilon threshold ($\epsilon_{IRL}$). In that case, the algorithm's convergence process needs more time to meet the threshold; however, the agent feature expectation vector ($\mu(\pi_i)$) is closer to the expert feature expectation vector ($\bar\mu(\pi_E)$). Therefore, the threshold value can be determined based on the user's expectation for the learning and optimization process of the reward function's weights and the optimal policy. Any changes in the epsilon threshold ($\epsilon_{IRL}$) may change the results shown in Fig.~\ref{fig:Hyper_distance}, and more iterations may be required to meet the criteria. 

\begin{figure}[t!]
	\centering
	\includegraphics[width=0.8\columnwidth]{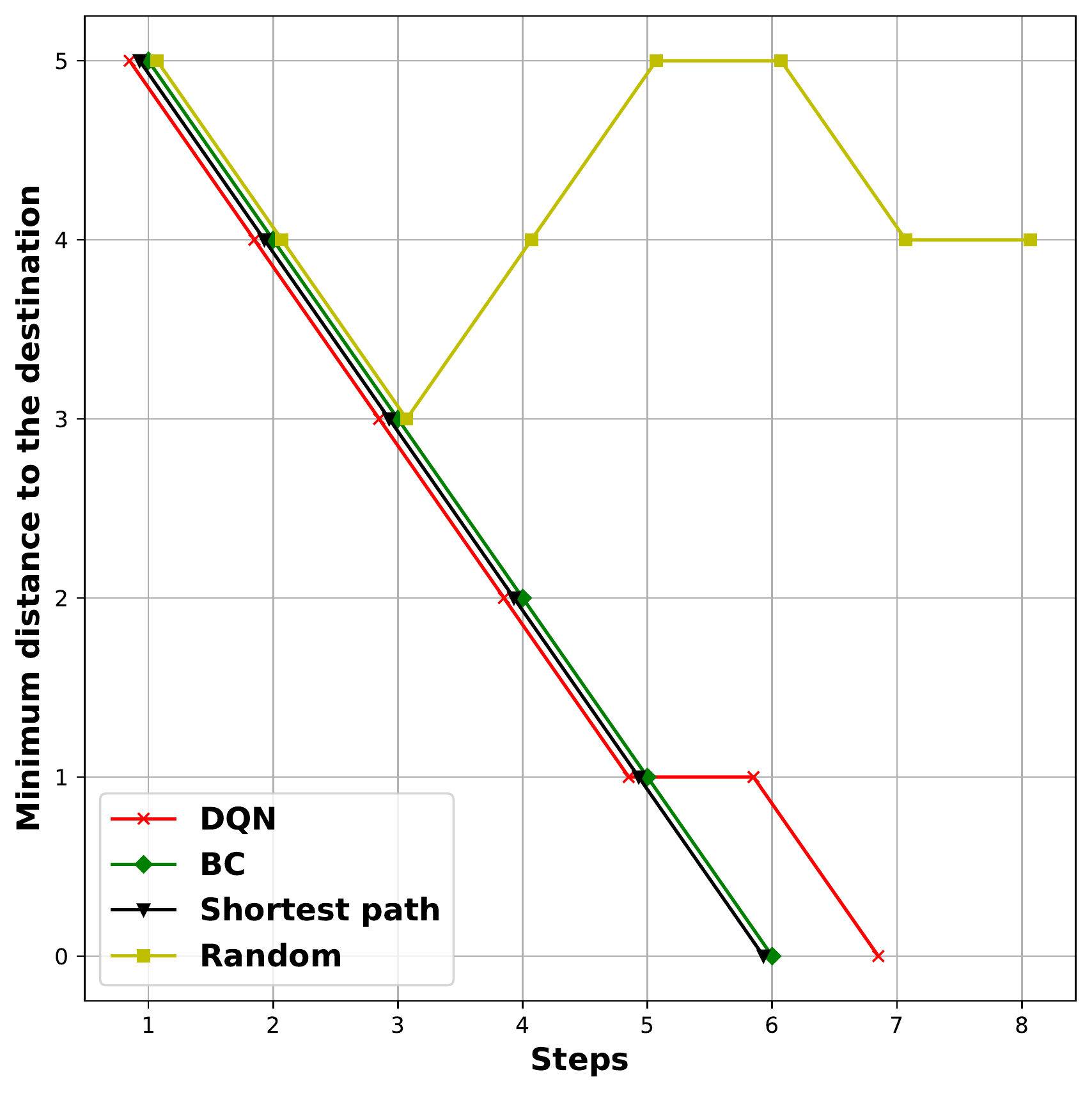}
	\caption{The distance between the last location where the UAV stops its task and the center of the destination cell}
    \label{fig:Distance_sample_steps_7}
\end{figure}

Fig.~\ref{fig:Distance_sample_steps_7} shows the distance between the destination cell (BS$_{24}$) and the cell where the agent stops or finishes its tasks. Both the behavioral cloning and shortest path visit the
destination cell at the 6th step. DQN visits the destination cell at the $7^{th}$ step. The random policy never finds the destination cell because of battery life and flight time limitations.

\subsection{Inverse RL and Behavioral Cloning in Unseen States by the Expert}
\label{subsec:sim_IRL_BC_error}

Another scenario is studied in this section to compare the performance of the Q-learning, DQN, and BC in a situation where the UAV is placed in a cell that an expert has never seen or experienced. Since we assumed that the expert is only available for a few trajectories or it is costly to have on-demand access to the expert, it is not possible to ask the expert to experience the new state. Hence, approaches like DAGGER are not practical in this situation. In this case, an environmental error such as wind is applied to the drone and placed the drone in the adjacent cell ($BS_5$) as the initial cell or the cell of the source. This $BS_5$ has never been seen by an expert before, and there is no data available for this point. 

\begin{figure*}
    \centering
    
    \begin{subfigure}{0.32\linewidth}
        \centering
        \includegraphics[width=\textwidth]{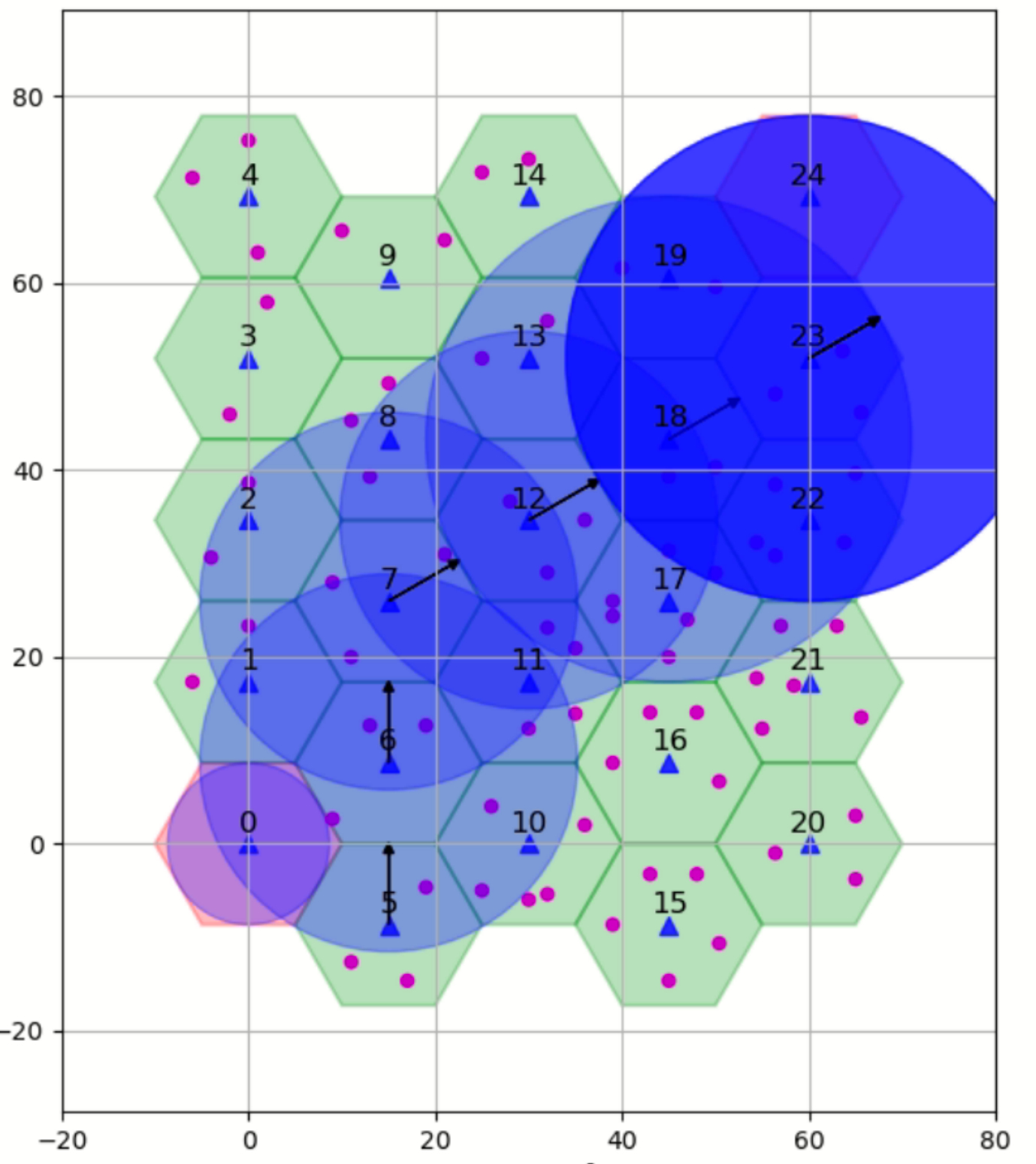}
        \caption{Path of the behavioral cloning with the applied error}
        \label{subfig:path_error_bc}
    \end{subfigure}
    \hfill
    \begin{subfigure}{0.32\linewidth}
        \centering
        \includegraphics[width=\textwidth]{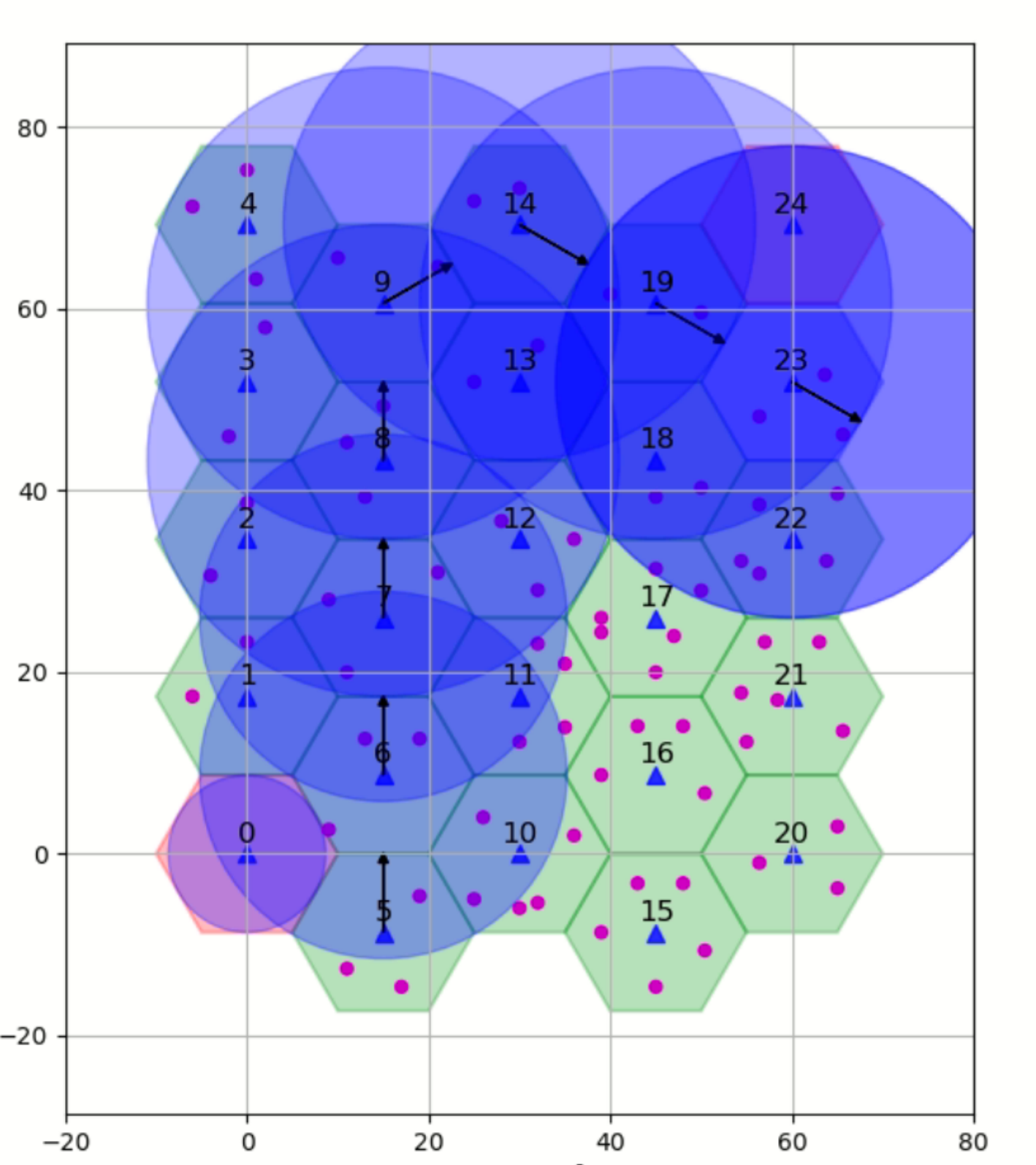}
        \caption{Path of the Q-learning with the applied error}
        \label{subfig:path_error_qlearning}
    \end{subfigure}
    \hfill
    \begin{subfigure}{0.32\linewidth}
        \centering
        \includegraphics[width=\textwidth]{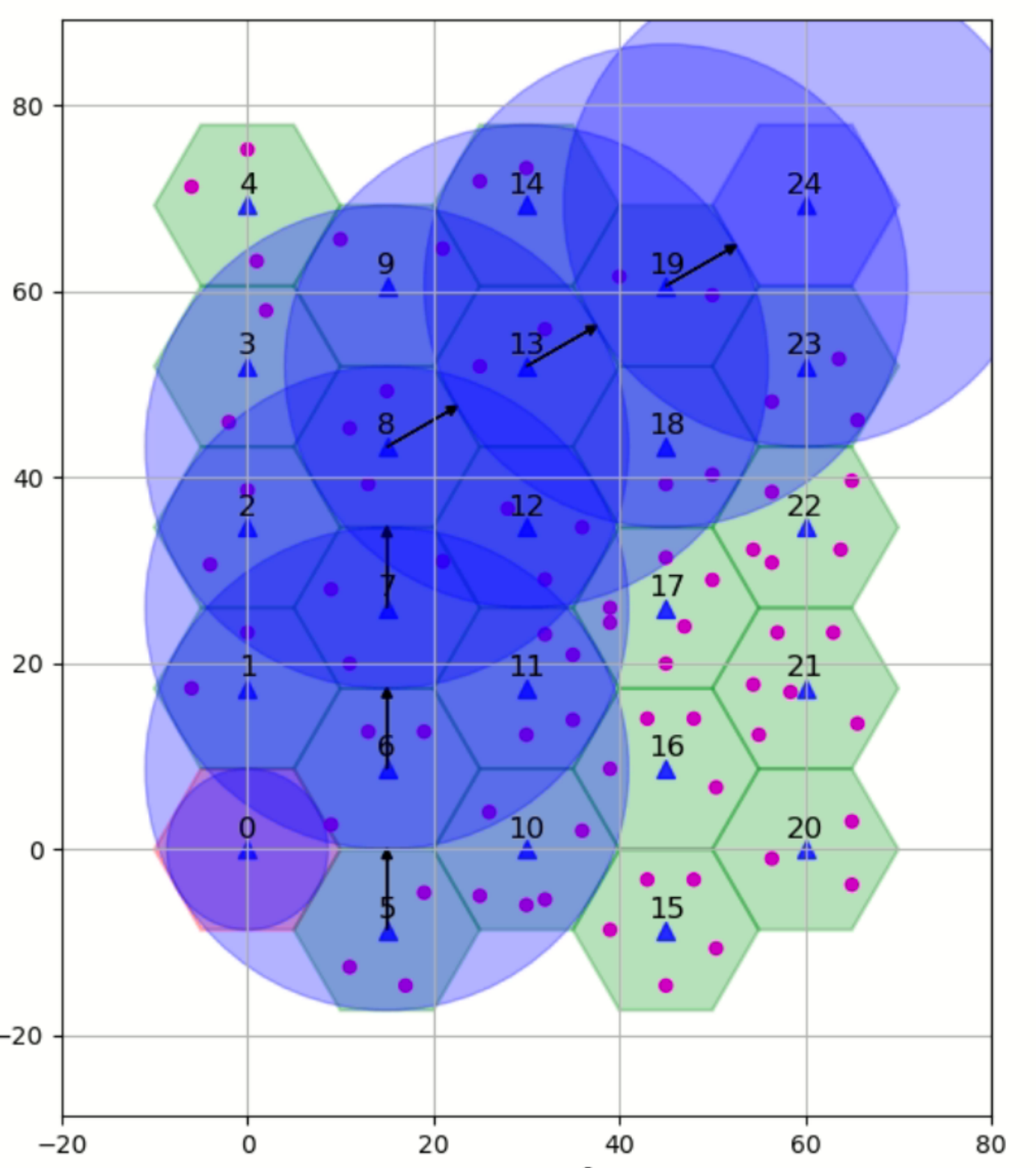}
        \caption{Path of the deep Q-network with the applied error}
        \label{subfig:path_error_dqn}
    \end{subfigure}

    \caption{Running Q-learning, DQN, and behavioral cloning on the defined scenario with an environmental error where wind moves the UAV to from cell(0) to the adjacent cell.}
    \label{fig:paths_error}
\end{figure*}

Fig.~\ref{fig:paths_error} shows the agent's trajectories, states, and actions for the three approaches mentioned. Based on Fig.~\ref{subfig:path_error_bc} the UAV utilizing the BC approach starts its location from BS$_5$, and it chooses the path with a higher density of UE, and it never reaches the destination cell (BS$_{24}$) to finish its task. Also, Q-learning in Fig.~\ref{subfig:path_error_qlearning} utilizes the Q-learning approach and tried to avoid the high UE density area but it fails to reach the destination because of a complicated dynamic wireless situation. However, DQN found the destination cell and tried to avoid the high UE density area considering the shortest path as well in Fig.~\ref{subfig:path_error_dqn}.

\begin{figure*}
    \centering
    
    \begin{subfigure}{0.32\linewidth}
        \centering
        \includegraphics[width=\textwidth]{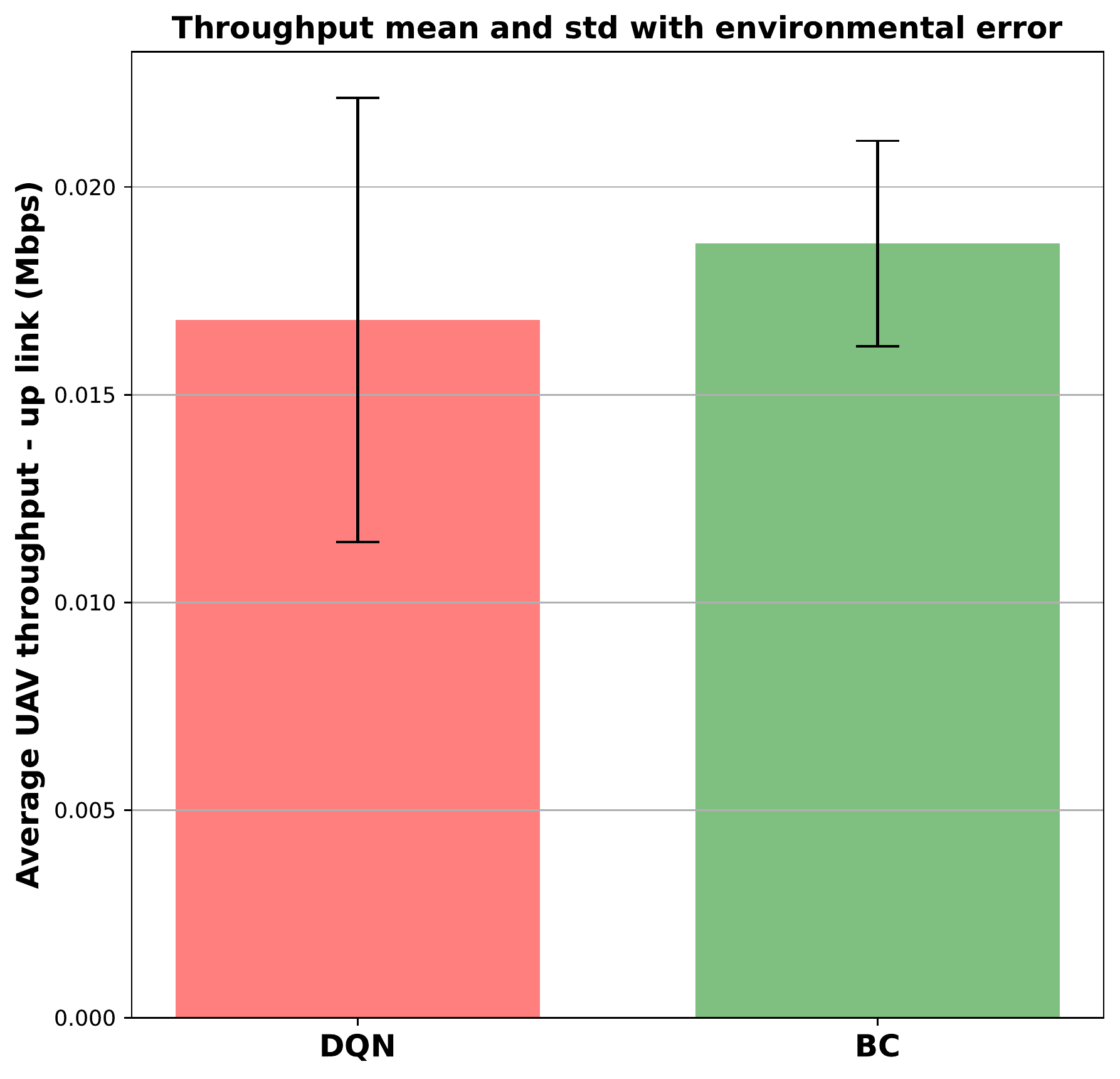}
        \caption{Throughput mean and standard deviation for the UAV UL}
        \label{subfig:throughput_error}
    \end{subfigure}
    \hfill
    \begin{subfigure}{0.32\linewidth}
        \centering
        \includegraphics[width=\textwidth]{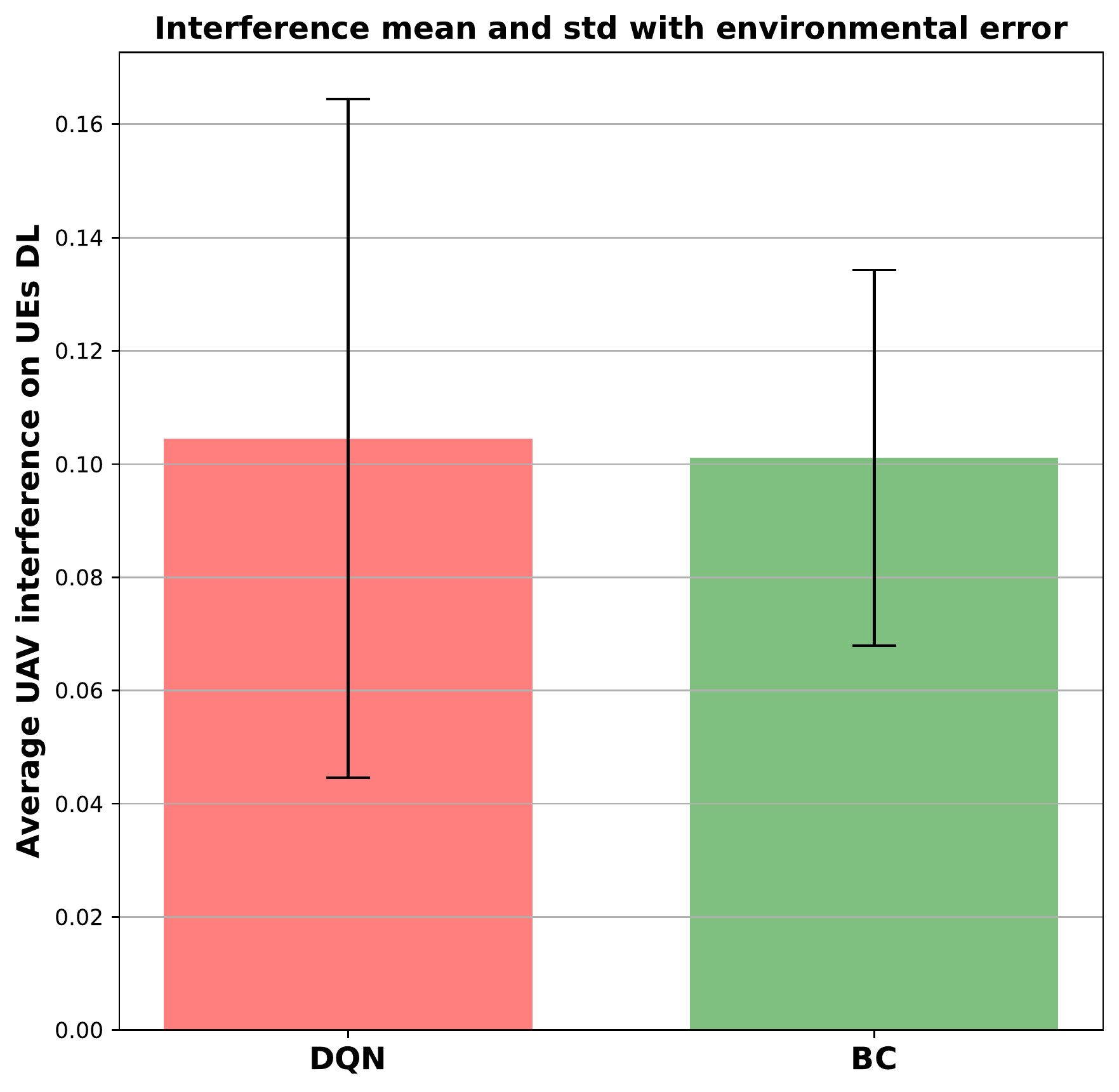}
        \caption{Interference mean and standard deviation on UEs DLs}
        \label{subfig:interference_error}
    \end{subfigure}
    \hfill
    \begin{subfigure}{0.32\linewidth}
        \centering
        \includegraphics[width=\textwidth]{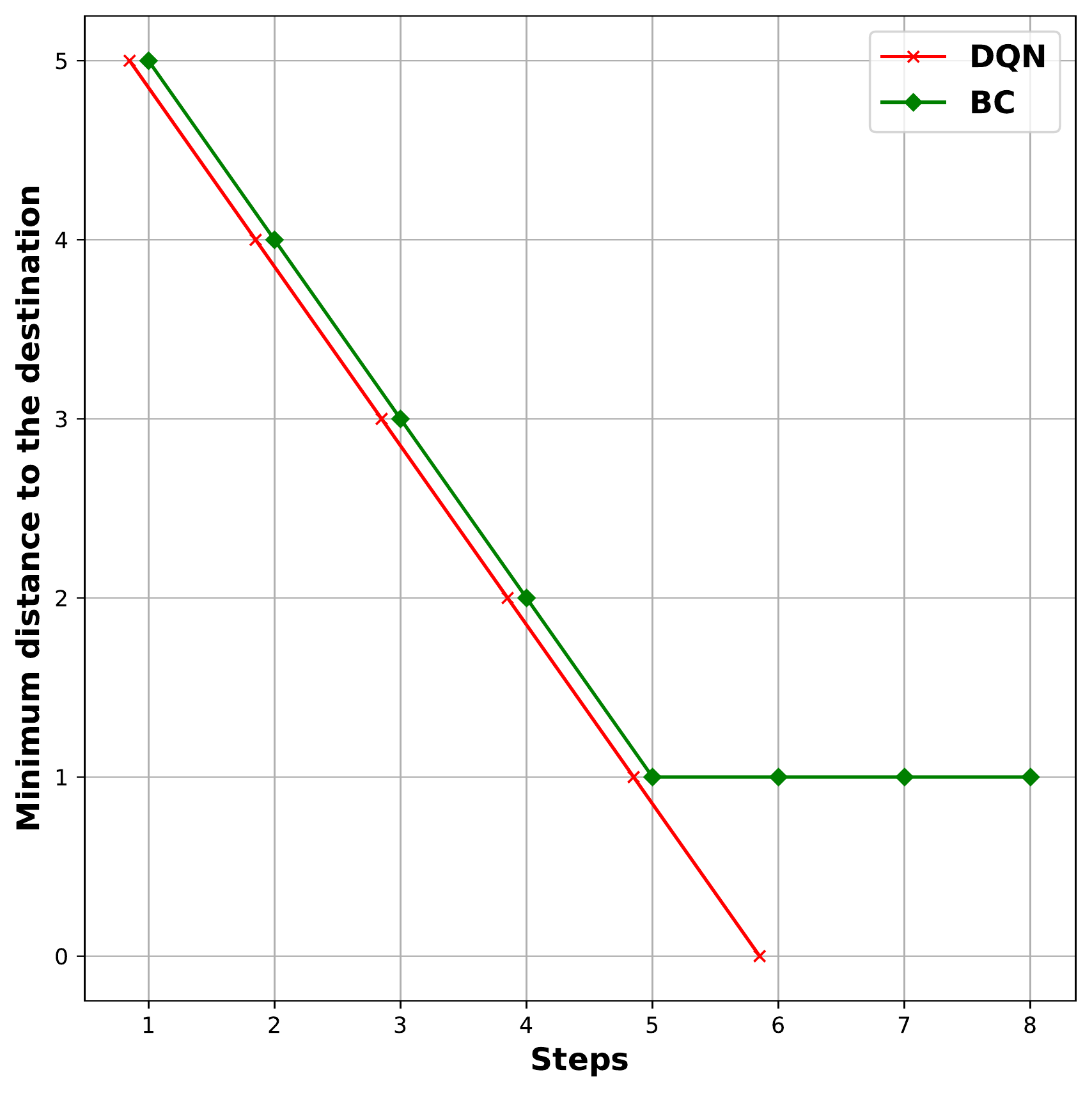}
        \caption{Distance between the UAV and the destination point}
        \label{subfig:distance_error}
    \end{subfigure}

    \caption{Performance of DQN, and BC with the applied error.}
    \label{fig:performance_error}
\end{figure*}

The performance evaluation of throughput, interference and the distance between the drone and the destination cell is depicted in Fig.~\ref{fig:performance_error}. DQN approach has better average and standard deviation for the throughput in Fig.~\ref{subfig:throughput_error}. The reason is that the BC chooses power values for its path, which is incorrect, since the path has never been experienced before, and the data for that was not available in the dataset. 
Fig.~\ref{subfig:interference_error} demonstrates that DQN has better interference management in the case that any errors happen to the system. Also, Fig.~\ref{subfig:distance_error} shows that DQN reached the destination and finished the task at the $6^{th}$ step; however, the BC was not able to finish the task by reaching the destination cell.

This evaluation shows that apprenticeship learning using inverse deep RL performs better compared to imitation learning (behavioral cloning) for the cases where the visited state by the agent was not experienced by the expert.

One future direction for this work is to study a more practical scenario for modeling the antenna gains as the antennas of the ground BSs are usually down tilted and optimized to increase the capacity for ground UEs and minimize the inter-cell interference effect to the ground UEs.
Hence, the cellular-connected UAVs may be serviced by the sidelobes of the terrestrial BSs. 
In this case, the UAV may connect to BSs which are located far away, and the channel state information shows non-trivial behavior. Studies such as \cite{chowdhury20203} and \cite{chowdhury2021taxonomy} considered the antenna behavior for trajectory optimization and observation in both simulation and experiments. Considering this challenge can propose the future direction regarding the system model and methodology in this paper.


\section{Conclusions}
\label{sec:Conclusion}

The emerging applications of unmanned aerial vehicles call for new reliable communication technologies to facilitate a low-delay high transmission rate communication. Cellular networks, in particular 5G and beyond, offer various advantages for drones including reliability, wide coverage, and security. However, noting the fast-growing number of utilized drones as aerial cellular users may raise several challenges such as interference to terrestrial users and base stations. 
In this paper, a novel interference-aware joint path planning and power allocation scheme are proposed for an autonomous cellular-connected unmanned aerial vehicle. 

The UAV's object is to take a path from its initial point to its destination subject to minimizing the interference imposed on the terrestrial User Equipment (UEs) and maximizing its uplink throughput. To solve this problem, we proposed apprenticeship learning via inverse RL using both Q-learning and deep Q-network solutions to find the optimal path planning with power allocation using expert knowledge. Another imitation learning method called behavioral cloning is performed using the decision tree supervised learning to compare the proposed method for the sake of comparison. In the numerical and simulation results, the apprenticeship learning via IRL performs close to the expert and the BC approach. Also, the user can define the threshold to determine how close the agent should follow the desired expert behavior.
Moreover, we showed that inverse RL performs better in cases where an error exists in the system or the supervised data for the visited state was never collected by the expert. 




\section*{Acknowledgment}
This material is based upon work supported by the Air Force Office of Scientific Research under award number FA9550-20-1-0090 and the National Science Foundation under Grant Numbers CNS-2232048 and ECCS 2030047.

 \bibliographystyle{elsarticle-num} 
 \bibliography{main}

\begin{thebibliography}{10}
\expandafter\ifx\csname url\endcsname\relax
  \def\url#1{\texttt{#1}}\fi
\expandafter\ifx\csname urlprefix\endcsname\relax\def\urlprefix{URL }\fi
\expandafter\ifx\csname href\endcsname\relax
  \def\href#1#2{#2} \def\path#1{#1}\fi

\bibitem{andreeva2020supporting}
A.~Andreeva-Mori, D.~Kubo, K.~Kobayashi, Y.~Okuno, J.~R. Homola, M.~Johnson,
  P.~H. Kopardekar, Supporting disaster relief operations through utm:
  Operational concept and flight tests of unmanned and manned vehicles at a
  disaster drill, in: AIAA Scitech 2020 Forum, 2020, p. 2202.

\bibitem{pandey2020adaptive}
A.~Pandey, P.~K. Shukla, R.~Agrawal, An adaptive flying ad-hoc network
  {{(FANET)}} for disaster response operations to improve quality of service
  (qos), Modern Physics Letters B 34~(10) (2020) 2050010.

\bibitem{javed2023uav}
S.~Javed, A.~Hassan, R.~Ahmad, W.~Ahmed, M.~M. Alam, J.~J. Rodrigues, Uav
  trajectory planning for disaster scenarios, Vehicular Communications (2023)
  100568.

\bibitem{AfghahINFOCOM}
F.~{Afghah}, A.~{Razi}, J.~{Chakareski}, J.~{Ashdown}, Wildfire monitoring in
  remote areas using autonomous unmanned aerial vehicles, in: IEEE INFOCOM 2019
  - IEEE Conference on Computer Communications Workshops (INFOCOM WKSHPS),
  2019, pp. 835--840.
\newblock \href {https://doi.org/10.1109/INFCOMW.2019.8845309}
  {\path{doi:10.1109/INFCOMW.2019.8845309}}.

\bibitem{shamsoshoara2021aerial}
A.~Shamsoshoara, F.~Afghah, A.~Razi, L.~Zheng, P.~Z. Ful{\'e}, E.~Blasch,
  Aerial imagery pile burn detection using deep learning: the flame dataset,
  Computer Networks (2021) 108001.

\bibitem{keshavarz2020real}
M.~Keshavarz, A.~Shamsoshoara, F.~Afghah, J.~Ashdown, A real-time framework for
  trust monitoring in a network of unmanned aerial vehicles, in: IEEE INFOCOM
  2020-IEEE Conference on Computer Communications Workshops (INFOCOM WKSHPS),
  IEEE, 2020, pp. 677--682.

\bibitem{FAAAeros18:online}
Faa aerospace forecasts,
  \url{https://www.faa.gov/data_research/aviation/aerospace_forecasts/},
  (Accessed on 03/06/2023).

\bibitem{forecast2020fiscal}
F.~Forecast, Fiscal years 2022-2042, Federal Aviation Administration (2022).

\bibitem{mozaffari2016unmanned}
M.~Mozaffari, W.~Saad, M.~Bennis, M.~Debbah, Unmanned aerial vehicle with
  underlaid device-to-device communications: Performance and tradeoffs, IEEE
  Transactions on Wireless Communications 15~(6) (2016) 3949--3963.

\bibitem{mozaffari2019tutorial}
M.~Mozaffari, W.~Saad, M.~Bennis, Y.-H. Nam, M.~Debbah, A tutorial on uavs for
  wireless networks: Applications, challenges, and open problems, IEEE
  Communications Surveys \& Tutorials 21~(3) (2019) 2334--2360.

\bibitem{Huang}
Q.~{Huang}, A.~{Razi}, F.~{Afghah}, P.~{Fule}, Wildfire spread modeling with
  aerial image processing, in: 2020 IEEE 21st International Symposium on "A
  World of Wireless, Mobile and Multimedia Networks" (WoWMoM), 2020, pp.
  335--340.
\newblock \href {https://doi.org/10.1109/WoWMoM49955.2020.00063}
  {\path{doi:10.1109/WoWMoM49955.2020.00063}}.

\bibitem{zeng2016wireless}
Y.~Zeng, R.~Zhang, T.~J. Lim, Wireless communications with unmanned aerial
  vehicles: Opportunities and challenges, IEEE Communications Magazine 54~(5)
  (2016) 36--42.

\bibitem{Nima_Infocom}
N.~Namvar, F.~Afghah, Heterogeneous airborne mmwave cells: Optimal placement
  for power-efficient maximum coverage, in: IEEE INFOCOM 2022 - IEEE Conference
  on Computer Communications Workshops (INFOCOM WKSHPS), 2022, pp. 1--6.
\newblock \href {https://doi.org/10.1109/INFOCOMWKSHPS54753.2022.9798023}
  {\path{doi:10.1109/INFOCOMWKSHPS54753.2022.9798023}}.

\bibitem{Nima_Asilomar}
N.~Namvar, F.~Afghah, Joint 3d placement and interference management for drone
  small cells, in: 2021 55th Asilomar Conference on Signals, Systems, and
  Computers, 2021, pp. 780--784.
\newblock \href {https://doi.org/10.1109/IEEECONF53345.2021.9723350}
  {\path{doi:10.1109/IEEECONF53345.2021.9723350}}.

\bibitem{3GPP_online}
3rd Generation Partnership Project~(3GPP), 3rd generation partnership project
  (3gpp), release 15, \url{https://www.3gpp.org/specifications/67-releases},
  (Accessed on 05/12/2020) (March 2020).

\bibitem{3GPP_online_subband}
3rd Generation Partnership Project~(3GPP), 3gpp ts 36.420 3gpp tsg ran evolved
  universal terrestrial radio access network (eutran), x2 general aspects and
  principles, version 11.0.0, release 11,
  \url{https://www.3gpp.org/specifications/67-releases}, (Accessed on
  15/17/2020) (October 2012).

\bibitem{rovira2022review}
A.~Rovira-Sugranes, A.~Razi, F.~Afghah, J.~Chakareski, A review of ai-enabled
  routing protocols for uav networks: Trends, challenges, and future outlook,
  Ad Hoc Networks 130 (2022) 102790.

\bibitem{9839122}
F.~Lotfi, O.~Semiari, W.~Saad, Semantic-aware collaborative deep reinforcement
  learning over wireless cellular networks, in: ICC 2022 - IEEE International
  Conference on Communications, 2022, pp. 5256--5261.
\newblock \href {https://doi.org/10.1109/ICC45855.2022.9839122}
  {\path{doi:10.1109/ICC45855.2022.9839122}}.

\bibitem{lahmeri2021artificial}
M.-A. Lahmeri, M.~A. Kishk, M.-S. Alouini, Artificial intelligence for
  uav-enabled wireless networks: A survey, IEEE Open Journal of the
  Communications Society 2 (2021) 1015--1040.

\bibitem{huang2021massive}
Y.~Huang, Q.~Wu, R.~Lu, X.~Peng, R.~Zhang, Massive mimo for cellular-connected
  uav: Challenges and promising solutions, IEEE Communications Magazine 59~(2)
  (2021) 84--90.

\bibitem{lotfi2022}
F.~{Lotfi}, O.~{Semiari}, F.~{Afghah}, Evolutionary deep reinforcement learning
  for dynamic slice management in {O-RAN}, arXiv preprint arXiv:2208.14394
  (2022).

\bibitem{huo20175g}
Y.~Huo, X.~Dong, W.~Xu, 5g cellular user equipment: From theory to practical
  hardware design, IEEE Access 5 (2017) 13992--14010.

\bibitem{checko2014cloud}
A.~Checko, H.~L. Christiansen, Y.~Yan, L.~Scolari, G.~Kardaras, M.~S. Berger,
  L.~Dittmann, Cloud ran for mobile networks—a technology overview, IEEE
  Communications surveys \& tutorials 17~(1) (2014) 405--426.

\bibitem{Mohammed_ICC}
M.~Gharib, S.~Nandadapu, F.~Afghah, An exhaustive study of using commercial lte
  network for uav communication in rural areas (2021).
\newblock \href {http://arxiv.org/abs/2105.03778} {\path{arXiv:2105.03778}}.

\bibitem{9448665}
F.~Lotfi, O.~Semiari, Performance analysis and optimization of uplink cellular
  networks with flexible frame structure, in: 2021 IEEE 93rd Vehicular
  Technology Conference (VTC2021-Spring), 2021, pp. 1--5.
\newblock \href {https://doi.org/10.1109/VTC2021-Spring51267.2021.9448665}
  {\path{doi:10.1109/VTC2021-Spring51267.2021.9448665}}.

\bibitem{3GPP_16}
3rd Generation Partnership Project~(3GPP), 5g;unmanned aerial system (uas)
  support in 3gpp, 3gpp ts 22.125 version 16.3.0 release 16, technical
  specification, \url{https://www.3gpp.org/specifications/67-releases}
  (November 2020).

\bibitem{3GPP_17}
3rd Generation Partnership Project~(3GPP), Universal mobile telecommunications
  system (umts); lte; 5g; t8 reference point for northbound apis, 3gpp ts
  29.122 version 17.5.0 release 17, technical specification,
  \url{https://www.3gpp.org/specifications/67-releases} (May 2022).

\bibitem{mishra2020survey}
D.~Mishra, E.~Natalizio, A survey on cellular-connected uavs: Design
  challenges, enabling 5g/b5g innovations, and experimental advancements,
  Computer Networks 182 (2020) 107451.

\bibitem{zeng2021simultaneous}
Y.~Zeng, X.~Xu, S.~Jin, R.~Zhang, Simultaneous navigation and radio mapping for
  cellular-connected uav with deep reinforcement learning, IEEE Transactions on
  Wireless Communications (2021).

\bibitem{mei2019cellular}
W.~Mei, Q.~Wu, R.~Zhang, Cellular-connected uav: Uplink association, power
  control and interference coordination, IEEE Transactions on Wireless
  Communications 18~(11) (2019) 5380--5393.

\bibitem{zeng2019path}
Y.~Zeng, X.~Xu, Path design for cellular-connected uav with reinforcement
  learning, in: 2019 IEEE Global Communications Conference (GLOBECOM), IEEE,
  2019, pp. 1--6.

\bibitem{shamsoshoara2019distributed}
A.~Shamsoshoara, M.~Khaledi, F.~Afghah, A.~Razi, J.~Ashdown, Distributed
  cooperative spectrum sharing in uav networks using multi-agent reinforcement
  learning, in: 2019 16th IEEE Annual Consumer Communications \& Networking
  Conference (CCNC), IEEE, 2019, pp. 1--6.

\bibitem{shamsoshoara2019solution}
A.~Shamsoshoara, M.~Khaledi, F.~Afghah, A.~Razi, J.~Ashdown, K.~Turck, A
  solution for dynamic spectrum management in mission-critical uav networks,
  in: 2019 16th Annual IEEE International Conference on Sensing, Communication,
  and Networking (SECON), IEEE, 2019, pp. 1--6.

\bibitem{shamsoshoara2020autonomous}
A.~Shamsoshoara, F.~Afghah, A.~Razi, S.~Mousavi, J.~Ashdown, K.~Turk, An
  autonomous spectrum management scheme for unmanned aerial vehicle networks in
  disaster relief operations, IEEE Access 8 (2020) 58064--58079.

\bibitem{challita2019interference}
U.~Challita, W.~Saad, C.~Bettstetter, Interference management for
  cellular-connected uavs: A deep reinforcement learning approach, IEEE
  Transactions on Wireless Communications 18~(4) (2019) 2125--2140.

\bibitem{riley2010ns}
G.~F. Riley, T.~R. Henderson, The ns-3 network simulator, in: Modeling and
  tools for network simulation, Springer, 2010, pp. 15--34.

\bibitem{gomez2016srslte}
I.~Gomez-Miguelez, A.~Garcia-Saavedra, P.~D. Sutton, P.~Serrano, C.~Cano, D.~J.
  Leith, srslte: An open-source platform for lte evolution and experimentation,
  in: Proceedings of the Tenth ACM International Workshop on Wireless Network
  Testbeds, Experimental Evaluation, and Characterization, 2016, pp. 25--32.

\bibitem{nikaein2014openairinterface}
N.~Nikaein, M.~K. Marina, S.~Manickam, A.~Dawson, R.~Knopp, C.~Bonnet,
  Openairinterface: A flexible platform for 5g research, ACM SIGCOMM Computer
  Communication Review 44~(5) (2014) 33--38.

\bibitem{shamsoshoara2021uav}
A.~Shamsoshoara, F.~Afghah, E.~Blasch, J.~Ashdown, M.~Bennis, Uav-assisted
  communication in remote disaster areas using imitation learning, IEEE Open
  Journal of the Communications Society (2021).

\bibitem{abbeel2004apprenticeship}
P.~Abbeel, A.~Y. Ng, Apprenticeship learning via inverse reinforcement
  learning, in: Proceedings of the twenty-first international conference on
  Machine learning, 2004, p.~1.

\bibitem{wu2018joint}
Q.~Wu, Y.~Zeng, R.~Zhang, Joint trajectory and communication design for
  multi-uav enabled wireless networks, IEEE Transactions on Wireless
  Communications 17~(3) (2018) 2109--2121.

\bibitem{al2014optimal}
A.~Al-Hourani, S.~Kandeepan, S.~Lardner, Optimal {LAP} altitude for maximum
  coverage, IEEE Wireless Communications Letters 3~(6) (2014) 569--572.

\bibitem{bain1995framework}
M.~Bain, C.~Sammut, A framework for behavioural cloning., in: Machine
  Intelligence 15, 1995, pp. 103--129.

\bibitem{osa2018algorithmic}
T.~Osa, J.~Pajarinen, G.~Neumann, J.~A. Bagnell, P.~Abbeel, J.~Peters, et~al.,
  An algorithmic perspective on imitation learning, Foundations and
  Trends{\textregistered} in Robotics 7~(1-2) (2018) 1--179.

\bibitem{adams2022survey}
S.~Adams, T.~Cody, P.~A. Beling, A survey of inverse reinforcement learning,
  Artificial Intelligence Review (2022) 1--40.

\bibitem{ross2011reduction}
S.~Ross, G.~Gordon, D.~Bagnell, A reduction of imitation learning and
  structured prediction to no-regret online learning, in: Proceedings of the
  fourteenth international conference on artificial intelligence and
  statistics, JMLR Workshop and Conference Proceedings, 2011, pp. 627--635.

\bibitem{ng2000algorithms}
A.~Y. Ng, S.~J. Russell, et~al., Algorithms for inverse reinforcement
  learning., in: Icml, Vol.~1, 2000, p.~2.

\bibitem{ratliff2006maximum}
N.~D. Ratliff, J.~A. Bagnell, M.~A. Zinkevich, Maximum margin planning, in:
  Proceedings of the 23rd international conference on Machine learning, 2006,
  pp. 729--736.

\bibitem{wulfmeier2015maximum}
M.~Wulfmeier, P.~Ondruska, I.~Posner, Maximum entropy deep inverse
  reinforcement learning, arXiv preprint arXiv:1507.04888 (2015).

\bibitem{zhou2017infinite}
Z.~Zhou, M.~Bloem, N.~Bambos, Infinite time horizon maximum causal entropy
  inverse reinforcement learning, IEEE Transactions on Automatic Control 63~(9)
  (2017) 2787--2802.

\bibitem{diamond2016cvxpy}
S.~Diamond, S.~Boyd, {CVXPY}: {A} {P}ython-embedded modeling language for
  convex optimization, Journal of Machine Learning Research 17~(83) (2016)
  1--5.

\bibitem{Home_CVX87:online}
Cvxopt, convex optimization for python, \url{https://cvxopt.org/}, (Accessed on
  04/07/2021).

\bibitem{sutton2018reinforcement}
R.~S. Sutton, A.~G. Barto, Reinforcement learning: An introduction, MIT press,
  2018.

\bibitem{levine2020offline}
S.~Levine, A.~Kumar, G.~Tucker, J.~Fu, Offline reinforcement learning:
  Tutorial, review, and perspectives on open problems, arXiv preprint
  arXiv:2005.01643 (2020).

\bibitem{shoham2003multi}
Y.~Shoham, R.~Powers, T.~Grenager, Multi-agent reinforcement learning: a
  critical survey, Tech. rep., Technical report, Stanford University (2003).

\bibitem{busoniu2006multi}
L.~Busoniu, R.~Babuska, B.~De~Schutter, Multi-agent reinforcement learning: A
  survey, in: 2006 9th International Conference on Control, Automation,
  Robotics and Vision, IEEE, 2006, pp. 1--6.

\bibitem{li2017convergence}
Y.~Li, Y.~Yuan, Convergence analysis of two-layer neural networks with relu
  activation, in: Advances in neural information processing systems, 2017, pp.
  597--607.

\bibitem{allen1971mean}
D.~M. Allen, Mean square error of prediction as a criterion for selecting
  variables, Technometrics 13~(3) (1971) 469--475.

\bibitem{kingma2014adam}
D.~P. Kingma, J.~Ba, Adam: A method for stochastic optimization, arXiv preprint
  arXiv:1412.6980 (2014).

\bibitem{torabi2018behavioral}
F.~Torabi, G.~Warnell, P.~Stone, Behavioral cloning from observation, arXiv
  preprint arXiv:1805.01954 (2018).

\bibitem{breiman1984classification}
L.~Breiman, J.~Friedman, C.~J. Stone, R.~A. Olshen, Classification and
  regression trees, CRC press, 1984.

\bibitem{scikit-learn}
F.~Pedregosa, G.~Varoquaux, A.~Gramfort, V.~Michel, B.~Thirion, O.~Grisel,
  M.~Blondel, P.~Prettenhofer, R.~Weiss, V.~Dubourg, J.~Vanderplas, A.~Passos,
  D.~Cournapeau, M.~Brucher, M.~Perrot, E.~Duchesnay, Scikit-learn: Machine
  learning in {P}ython, Journal of Machine Learning Research 12 (2011)
  2825--2830.

\bibitem{github:code_AL_IRL2021}
A.~Shamsoshoara, {Apprenticeship learning using deep inverse reinforcement
  learning code},
  \url{https://github.com/AlirezaShamsoshoara/Inverse-RL-Apprenticeship-learning-UAV-Communication}
  (2021).

\bibitem{youtube2021_inverserl}
A.~Shamsoshoara, {Apprenticeship learning via inverse RL for a
  cellular-connected UAV}, \url{https://youtu.be/FGAlHaTQ\_nc}, wireless
  Networking \& Information Processing (WINIP) LAB, accessed on 05/19/2021
  (2021).

\bibitem{chowdhury20203}
M.~M.~U. Chowdhury, S.~J. Maeng, E.~Bulut, I.~G{\"u}ven{\c{c}}, 3-d trajectory
  optimization in uav-assisted cellular networks considering antenna radiation
  pattern and backhaul constraint, IEEE Transactions on Aerospace and
  Electronic Systems 56~(5) (2020) 3735--3750.

\bibitem{chowdhury2021taxonomy}
M.~M.~U. Chowdhury, C.~K. Anjinappa, I.~Guvenc, M.~Sichitiu, O.~Ozdemir,
  U.~Bhattacherjee, R.~Dutta, V.~Marojevic, B.~Floyd, A taxonomy and survey on
  experimentation scenarios for aerial advanced wireless testbed platforms, in:
  2021 IEEE Aerospace Conference (50100), IEEE, 2021, pp. 1--20.

\end{thebibliography}

\end{document}